Sapienza
Università di Roma

# Training binary neural networks without floating point precision

Facoltà di Ingegneria dell'Informazione, Informatica e Statistica
Corso di Laurea Magistrale in Computer Science

Candidate
Federico Fontana
ID number 1744946

Thesis Advisor
Prof. Danilo Avola

Co-Advisors
Prof. Luigi Cinque
Dr. Romeo Lanzino

**Training binary neural networks without floating point precision**
Master's thesis. Sapienza – University of Rome

Author's email: scanafei@gmail.com

# Contents

# Chapter 1

# Introduction

This chapter is organized as follows. In section 1.1 the candidate's work, as well as the reasons that motivated it, are presented. In section 1.2 the binary neural network is defined. In section 1.3 the state of the art in relation to the work's topic, is summarized. Finally, in section 1.4, the contributions of the proposed work is highlighted and the structure of the thesis is also provided.

## 1.1 Aims and scope

AIs (Artificial Intelligence) are now an integral part of our daily life. People's lives are made easier by AI, which powers a variety of applications and services that let them accomplish things like connect with friends, send emails, and use ride-sharing services. Chatbots, smart automobiles, and IoT (Internet of Things) gadgets are the most recent breakthroughs in Artificial Intelligence. Artificial Intelligence is used in the healthcare, finance, logistics, and tourism industries to deliver a better experience.

In terms of environmental considerations, AI provides both a potential and a problem: we now have an electricity and carbon dilemma [1]. Video streaming, email, and social networking are examples of traditional data processing operations performed in data centers. Because AI must sift through a large amount of material until it learns to understand it – that is, until it is trained – it is more computationally intensive.

When compared to how humans learn, this teaching is ineffective. artificial neural networks, which are mathematical computations that replicate neurons in the human brain, are used in modern AI. Weight is a network parameter that describes the strength of each neuron's connection to its neighbors. The network starts with random weights and modifies them (through training where the model see samples) until the output agrees with the correct response to learn how to understand language. Bidirectional Encoder Representations from Transformers (BERT) was a recent model that utilised 3.3 billion words from English literature and Wikipedia articles. Furthermore, during training, BERT read this data set 40 times. By comparison, by the age of five, an ordinary youngster beginning to speak may have heard 45 million words, or 3,000 times fewer than BERT. [2].

In this work, efficient training methods are investigated. In specific, genetic algorithm, evolutionary algorithm and ad-hoc strategies are tested on a very efficient type of neural network: binary neural networks. The aim is to answer to the question: it is possible train binary neural networks without using floats, where a float is a data type composed of a number composed with 32 bits?

If a reliable way to train binary neural network without floats exists, then we can have network 32x smaller, 56x faster [3] and trainable with a lot less energy, memory and time.

## 1.2 Binary neural network overview

Deep learning has transformed the creation of intelligent systems in recent years and is now widely used in a variety of real-world applications. Deep learning processing is in significant demand in various computationally limited and energy-constrained systems, despite its various benefits and potentials. To improve deep learning skills, it's reasonable to look into game-changing technologies like binary neural networks. Them have recently made great development since they can be developed and incorporated on small, limited devices, saving a large amount of storage, computation cost, and energy consumption. A binary neural network is a kind of neural network in which all hidden layers' activations and weights are 1-bit numbers (except the input and output layers). Binarization is the process of reducing 32-bit information to 1-bit values. Binarization has the benefit of reducing matrix computing costs by employing XNOR and popcount operations, in addition to saving space both on memory and storage. [3] reported that binary neural networks can have 32 times memory saving and 58 times faster convolution operations than 32-bit convolutional neural networks. In full-precision convolutional neural networks, the basic operation can be expressed as:

$$\mathbf{Z} = \sigma(\mathbf{W} \otimes \mathbf{A}) \tag{1.1}$$

where $\sigma$ is the activation function, $\mathbf{W}$ are the weights, $\mathbf{A}$ are the activations and $\otimes$ is the convolution operator. Instead, in binary neural networks the operator can be reformulated as:

$$\mathbf{Z} = \sigma\left(Q_W(\mathbf{W}) \otimes Q_A(\mathbf{A})\right) = \sigma\left(\alpha\beta\left(\mathbf{B_W} \odot \mathbf{B_A}\right)\right) \tag{1.2}$$

with

$$Q_W(\mathbf{W}) = \alpha\mathbf{B_W}, \quad Q_A(\mathbf{A}) = \beta\mathbf{B_A} \tag{1.3}$$

where $\mathbf{B_W}$ and $\mathbf{B_A}$ are the tensor of binary weights (kernel) and binary activations, with the corresponding scalars $\alpha$ and $\beta$, and $\odot$ denotes the inner product for vectors with bitwise operation XNOR-bitcount. The XNOR-bitcount (Fig. 1.1 (a)) is composed by the XOR operator (Fig. 1.1 (b)) and the bitcount, the operation that takes as input a binary sequence, and return 0 if the sequence has more 0s than 1s, 1 otherwise. The *sign* function is widely

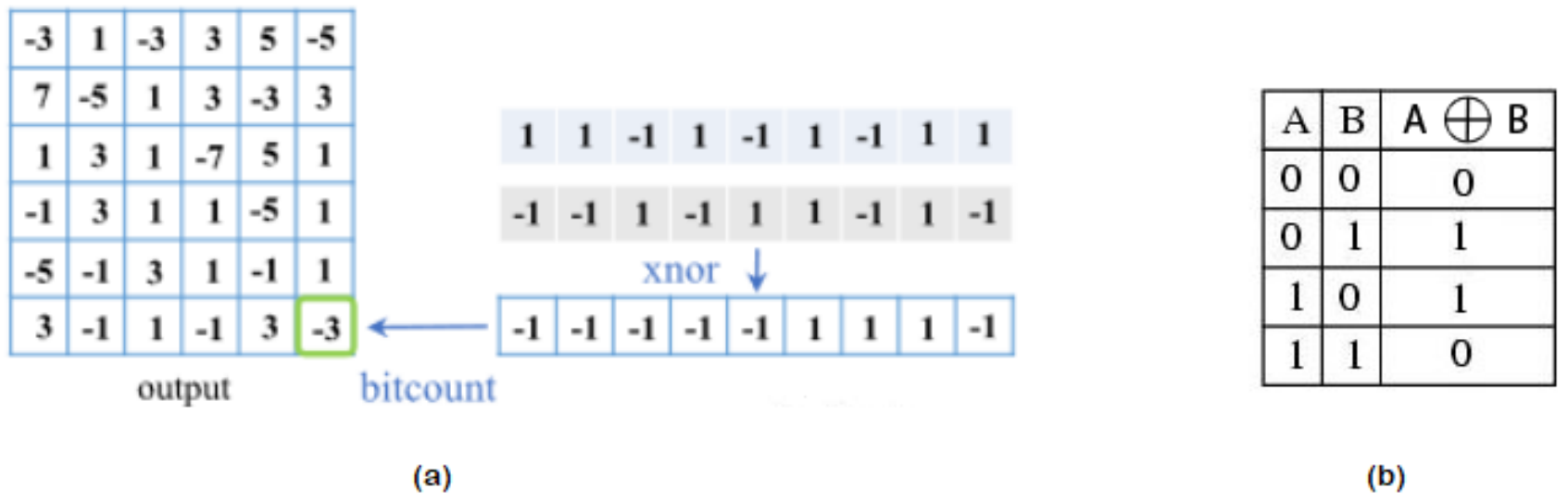

**Figure 1.1.** (a) XNOR-bitcount example and (b) XOR operator.

used for binarization (for obtaining $\mathbf{B_W}$ and $\mathbf{B_A}$).

$$sign(x) = \begin{cases} +1, & \text{if } x \geq 0 \\ -1, & \text{otherwise} \end{cases} \tag{1.4}$$

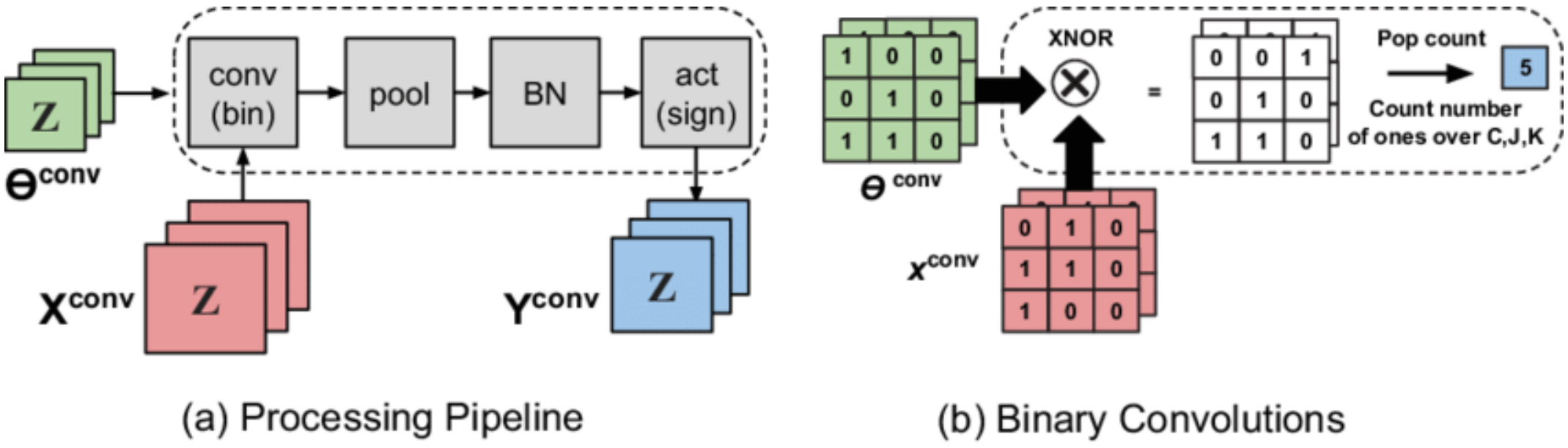

**Figure 1.2.** Processing pipeline (a) and binary convolution (b)[1].

## 1.3 State of the art

The current literature of binary neural networks rely on training with latent weights. In short, the network is trained with the conventional gradient descent and weights, and after the training all the weight are rounded to 0 and 1. The improvements papers in literature of binary neural networks can be divided in these categories:

- ***Quantization error minimization:*** the strategy is to lower the

[1]Image taken from the site: https://www.researchgate.net/figure/Binary-neural-networks$_{fig6_3}$40119036

information loss during *sign* function transformation from 32-bit values to 1-bit values. XNOR-Net [3] adds channel-wise scaling factors $\alpha$ and $\beta$ for activations and weights. DA-BNN [4], instead, design a data-adaptive method that can generate an adaptive amplitude based on spatial and channel attention.

- ***Loss function improvement:*** to close the accuracy gap from real-valued network, an ad-hoc distribution loss or special regularization can be added to the overall loss function. RBNN [5] use a mix of cross entropy and a regularized distribution

$$\mathrm{E}_T = \mathrm{E}_S + \lambda \mathrm{L}_{DR}$$

  where $\mathrm{E}_T$ is total loss, $\mathrm{E}_S$ is a cross-entropy loss, $\mathrm{L}_{DR}$ is the added special distribution loss or regularization and $\lambda$ is a balancing hyper-parameter.

- ***Gradient approximation:*** as the derivative result of sign function equals to zero, it leads weights fail to get updated in the back-propagation. Straight-through Estimator (STE) is one available method to approximate sign gradients. However, using STE fails to learn weights near the borders of $-1$ and $+1$, that greatly harms the updating ability of back propagation. SI-BNN [6] designs their gradient estimator with two trainable parameters on the top of STE.

- ***Network topology structure:*** neural architectural search and conditional computing made new architectures more accurate [7]. BENN [8] proposes to leverage ensemble binary neural network to improve prediction performance.

- ***Training strategy and tricks:*** different training scheme and tricks can also affect final accuracy of binary neural network. SQ-BWN [9] applied Stochastic Quantization (SQ) algorithm to gradually train and quantize binary neural networks to compensate the quantization error. Bi-Real-Net [10] initiated trainable parameters based on pre-train real value network and replaced ReLU activation function with hardtanh function. How to Train [11] replaced the ReLU activation function with PReLU function and explored that learning rate can have an influence on the accuracy of the final trained binary neural network.

To briefly recap, current state of the art approaches prefer to optimize the STE to train binary neural networks considering each component on the process, but not to explore training without latent weights.

## 1.4 Contributions and thesis outline

In this work the binary neural network training will be explored by making use of different methods that allow the network to learn but without using floats. To handle information loss, the system will borrow some techniques from other fields of study, e.g. genetic algorithm.

The main contributions of the proposed work, in relation to the current literature for training binary neural networks, can be summarized in two key points:

- Testing evolutionary and genetic algorithms to train binary neural networks using less time, energy and memory, including the introduction of an original training algorithm based on counting errors in bits;
- The introduction of novelties to make binary neural networks trainable with low-precision without losing much accuracy.

The rest of this thesis is structured as follows. In chapter 2 are showed the concepts under deep learning and modern efficient approaches. The chapter 3 will provide a theoretical explanation of gradient descent and its variations, as well as attempts in the literature to train neural networks using genetic algorithms. The attempts to train a full binary network with genetic algorithms are explained in chapter 4, as well as the final version of the proposed binary neural network and training strategy. Instead, chapter 5 contains the datasets used for the tests as well as the numerical results.

# Chapter 2

# Deep learning

This chapter is organized as follows. In section 2.1 there is an overview of machine learning tasks and learning paradigms. In section 2.2 the common approaches and networks are theoretically explored. In section 2.3 approaches and network more efficient in speed and memory are presented.

## 2.1 Introduction

Machine learning is increasingly ubiquitous in consumer items such as cameras and smartphones, and it powers many parts of modern life, from web searches to content filtering on social networks to suggestions on e-commerce websites. Systems based on such algorithms are used to recognize objects in photos, convert speech to text, match new items, postings or products to users' interests, and choose appropriate search results.

Deep learning methods are representation-learning methods with many levels of representation created by building simple but non-linear modules that change the representation at one level, beginning with the raw input, into a higher, slightly more abstract level. Higher layers of representation accentuate characteristics of the input that are significant for discrimination while suppressing irrelevant variations in classification tasks. When an image is represented as an array of pixel values, the features learned in the first layer of representation often represent the presence or absence of edges at specific orientations and places in the image.

### 2.1.1 Learning paradigms

#### 2.1.1.1 Supervised learning

Supervised learning is the most prevalent type of machine learning [12], whether deep or not. Assume we want to create a system that can identify whether an image contains a house, a car, a human, or a pet. We start by gathering a vast collection of images of houses, autos, people, and pets, all of which are labeled with their respective categories. An image is shown to the model and it generates an output in the form of a vector of scores, one for each category: we want the desired category to have the highest overall score, but this is highly unlikely before training.

The error (or distance) between the output scores and the desired pattern of scoring is calculated using an objective function; to reduce this inaccuracy, the machine adjusts its internal customizable parameters. These adjustable parameters, often known as weights, are real numbers that define the machine's input–output function. There might be hundreds of millions of these customizable weights in a typical deep-learning system, as well as hundreds of millions of annotated samples to train the machine with.

To appropriately change the weight vector, the learning algorithm creates a gradient vector that specifies how much the error would grow or reduce if the weight were increased by a little amount for each weight. Following that, the weight vector is modified in the opposite direction as the gradient vector. A common algorithm to handle this is the stochastic gradient descent 3.1.4 [13]. This entails displaying the input vector for a few examples, computing the outputs and errors, calculating the average gradient for those samples, and modifying the weights as needed. The procedure is repeated for a large number of small groups of samples from the training set until the objective function's average stops falling.

#### 2.1.1.2 Unsupervised learning

The training of models on raw and unlabeled training data is known as unsupervised learning. It's frequently used to spot patterns and trends in raw data, or to organize similar data into a set of categories. It's also a common method for better understanding datasets during the early exploration phase. When compared to supervised learning, unsupervised learning takes a more hands-off approach. Although a human will configure model's hyper-parameters like the number of cluster points, the model will process large amounts of data efficiently and without human intervention. As a result, this paradigm is well suited to answer queries concerning previously unknown patterns and relationships inside data.

However, because there is less human monitoring, the explainability of unsupervised learning should be given special attention. Since the vast bulk of data is unlabeled and unprocessed, unsupervised learning is a strong tool for gaining insight from data by grouping data based on similar attributes or analyzing datasets for underlying patterns. Due to the requirement for tagged data, supervised learning can be resource intensive.

### 2.1.2 Classification task

Machine learning is a discipline that studies algorithms that learn from examples. Classification is a task that necessitates the application of machine learning algorithms to learn how to assign a class label to problem domain instances: classifying emails as "spam" or "not spam" is an easy example to comprehend.

There are many different sorts of categorization jobs that can be encountered, and each one requires a distinct strategy.

#### 2.1.2.1 Binary classification

Classification problems with two class labels are referred to as binary classification. There are numerous measures that can be used to assess a classifier's or predictor's performance, and different sectors and disciplines have varied preferences for specific metrics based on their objectives; In most binary classification problems, one class represents the normal state and the other represents the aberrant state, like the in medical testing to determine if a patient has certain disease or not.

#### 2.1.2.2 Multi-class classification

The challenge of categorizing examples into one of three or more classes is known as multi-class or multinomial classification. An example of such classification type is image classification where the classes are more that two, like in the MNIST dataset (5.1.1) where the classes are the numbers from 0 to 9.

#### 2.1.2.3 Multi-label classification

In multi-label classification, each instance may be assigned many labels. Multi-label classification is an extension of multi-class classification in which there is no limit to how many classes an instance can be allocated to. An example of such classification type is text categorisation problem, where each document may belong to several predefined topics simultaneously.

### 2.1.3 Data augmentation

In data analysis, data augmentation refers to approaches for increasing the amount of data by adding slightly changed copies of current data or creating new synthetic data from existing data, both with hand-crafted algorithms or using deep neural networks. When training a machine learning model, this technique works as a regularizer and helps reduce overfitting.

By creating fresh and varied instances to train datasets, data augmentation can help improve the performance and results of machine learning models: a model works better and more correctly when the dataset is large and complete.

## 2.2 Common approaches

### 2.2.1 Architectures

#### 2.2.1.1 Perceptron

The simplest design that may be built is a perceptron, which is based on genuine neurons found in the human brain [14]. An actual neuron, as illustrated in Fig. 2.2.1.1, is made up of four basic components: dendrites, soma, axon, and synapses, which allow for the reception, processing, and transmission of electrical impulses. These signals are delivered and received between neurons via synapses and dendrites. It's worth noting that while summing all the incoming signals, a threshold must be met for transmission to take place. A perceptron models these four aspects of real neurons.

A perceptron can accept several inputs that are first multiplied by weights and then added together, as shown in the following equation:

$$net = b + \sum_{i=1}^{n} W_i X_i \tag{2.1}$$

where, $X_i$ represents the $i$-th input; $W_i$ is the weight associated to the $i$-th input and $b$ is the bias. The $net$ value is then used in conjuction with an activation function, generating an output for this specific perceptron. It is important to notice that various activation functions do exist, and their behaviour usually differs in how they treat the weighted sum of inputs.

Formally, the perceptron is an algorithm for learning a binary classifier termed a threshold function, which transfers its input $X$ to its output $Y$:

$$f(X) = \begin{cases} 1, & \text{if } b + \sum_{i=1}^{n} W_i X_i > 0 \\ 0, & \text{otherwise} \end{cases} \tag{2.2}$$

where $f(X)$ is the output function on the inputs of $W$, $X$ and $b$, that are defined as:

$$X = \{x_1, \quad x_2, \quad \dots, \quad x_n\} \tag{2.3}$$

$$W = \{w_0, \quad w_1, \quad \dots, \quad w_n\} \tag{2.4}$$

$b$ is the bias added. It is used in cases where the input and/or weight is 0 but the output required is greater than 0.

Historically, the perceptron was coupled with the sigmod activation function:

$$f(x) = \frac{1}{1 + e^{-x}} \tag{2.5}$$

We have to make use of some calculus to generate partial derivatives and update the weights. The partial derivatives are computed with the chain rule:

$$\frac{\delta error}{\delta W} = \frac{\delta error}{\delta y_{pred}} * \frac{\delta y_{pred}}{\delta comp} * \frac{\delta \text{ comp}}{\delta W} \tag{2.6}$$

$$\text{comp} = W * X + b \tag{2.7}$$

after the calculation, each derivative's weights are updated:

$$w = w - \alpha \frac{\delta \text{error}}{\delta w} \tag{2.8}$$

where $\alpha$ is the learning rate, a tuning parameter that determines the step size at each iteration while moving toward a minimum of a loss function.

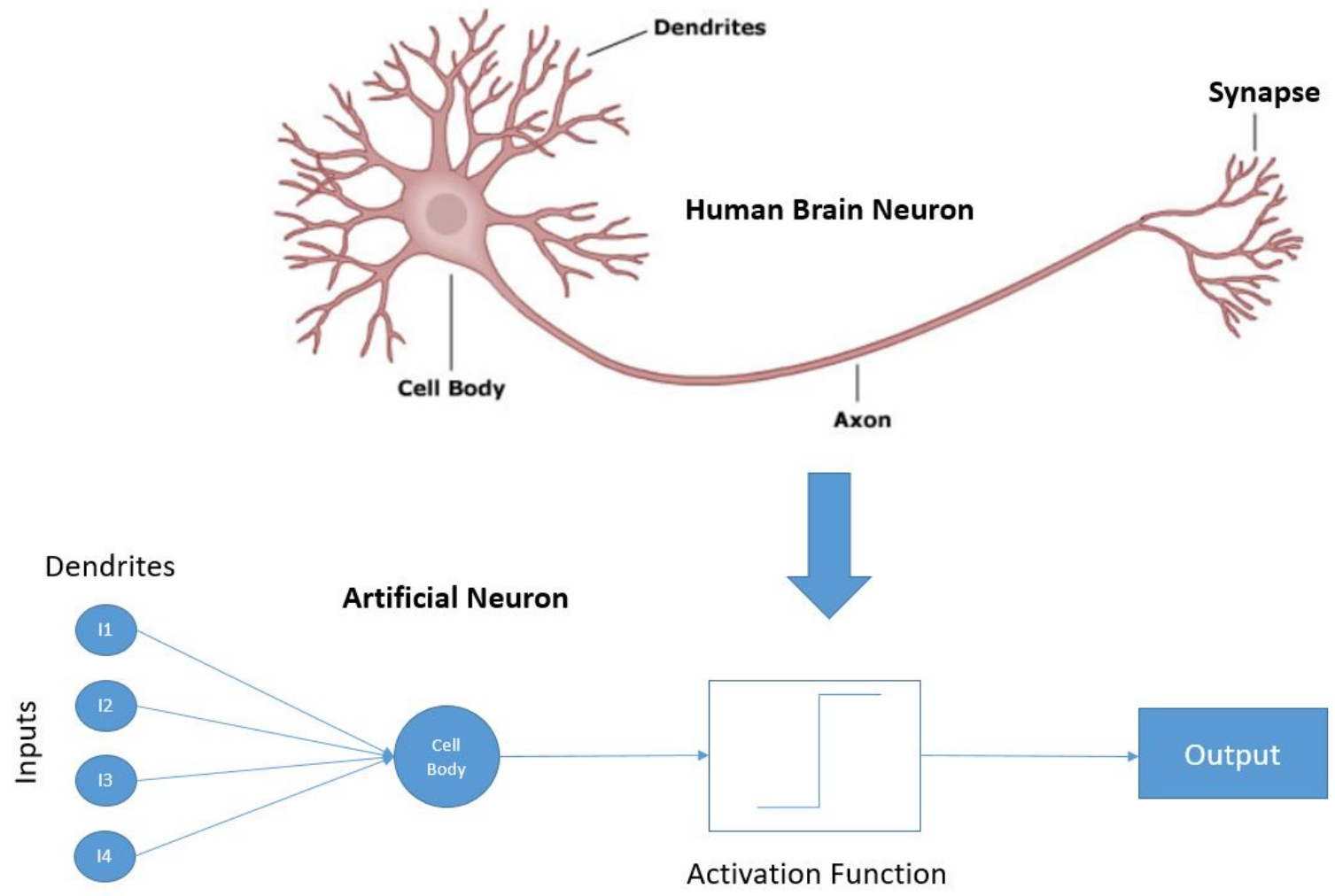

**Figure 2.1.** Neuron diagram with specified artificial neuron used in machine learning models.[1]

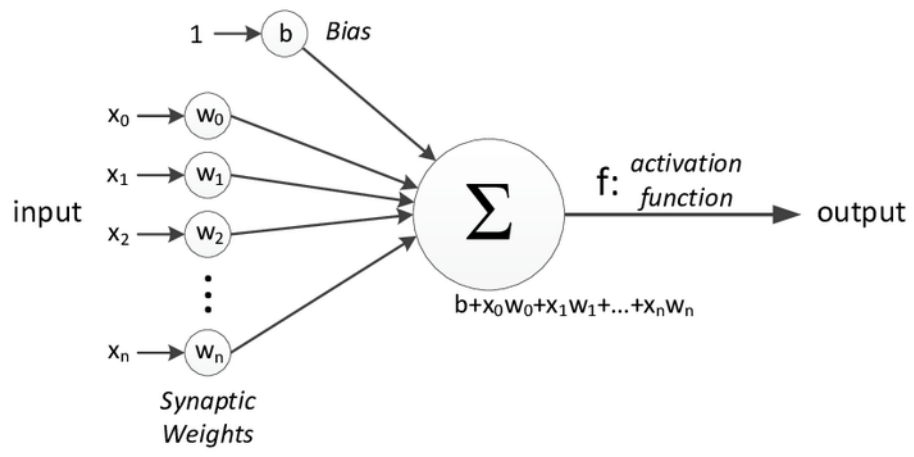

**Figure 2.2.** Neuron implementation example.

### 2.2.1.2 Multi-layer perceptron

A perceptron compensates for a neuron's poor estimates by handling nonlinearly separable inputs; the so-called multi-layer perceptron is a basic network architecture that solves such a problem [15]. A multi-layer perceptron is a type of feed-forward artificial neural network that is a fully connected neural network. The name multi-layer perceptron is ambiguous as it can be used to refer to any feed-forward network or it can refer to networks made up of many layers of perceptrons (with threshold activation), an example of it is showed in Fig. 2.3.

Multi-layer perceptron's weights are updated during training. In order to update the weights, a specific method is used, namely the back-propagation algorithm [16]. As previously stated, an optimization function is employed in this technique to select the optimal weights among the many layers. The function aims to reduce the loss between the network outputs and the real outcomes, which are supplied alongside the inputs.

[1]Image taken from the site: http://www.mplsvpn.info/2017/11/what-is-neuron-and-artificial-neuron-in.html

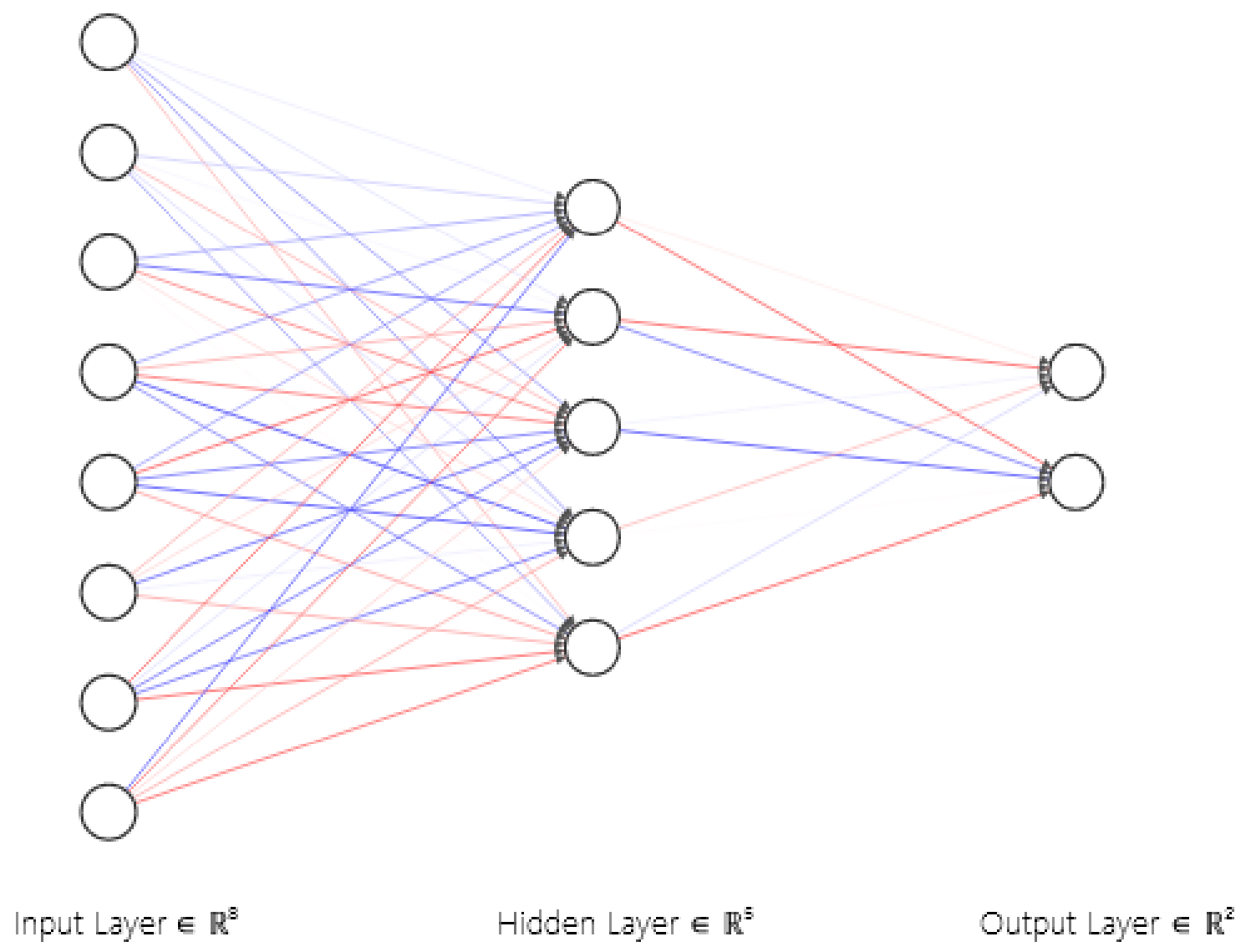

**Figure 2.3.** Multi-layer perceptron example with 8 input nodes, 5 nodes in the hidden layer, and an output layer composed by two nodes. the color represent the value of the weights.

The optimization function must, in the end, lower the error, and in the multi-class classification setting tipically is the cross-entropy loss calculated using the following equation:

$$E = -\frac{1}{|N|}\sum_{n=1}^{N}\sum_{i=1}^{M} d_{ni}\log y_{ni} \tag{2.9}$$

where $E$ is the value to be optimised, $N$ is a non-empty training set, $M$ are the classes, $y$ is the input, $d$ is the ground-truth.

A well-known network size issue known as "vanishing gradient" occurs when there is unit growth (i.e., more hidden layers are stacked together). The backpropagation algorithm employed in conjunction with a sigmoid activation function causes this issue, which results in the error not being transmitted appropriately (i.e., vanishing). Fortunately, there are answers to this issue: [17] frequently entail the use of a distinct activation function capable of preserving and propagating the error back. The ReLU activation function is an example of a common activation function used with the backpropagation algorithm, and it is also depicted in Fig. 2.4. Another solution for vanishing gradient is an architectural element in the network structure called skip connection [18].

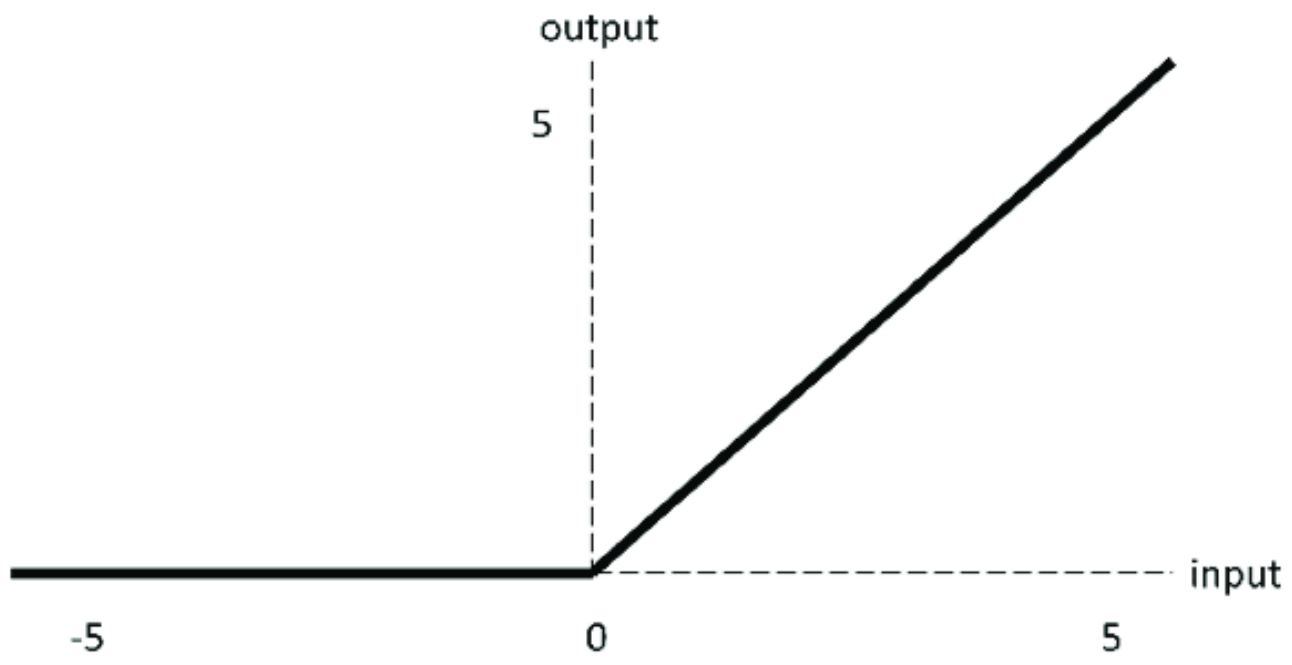

**Figure 2.4.** ReLU activation function.

#### 2.2.1.3 Convolutional neural networks

The convolutional neural network [19] is a common variant of the multi-layer perceptron that is based on the animal visual system, in which neurons only react to specified little sections of the total visual field, known as receptive fields. This common deep learning technique can detect patterns in presented data by analyzing sections of the sample in a manner comparable to the visual system of an animal. As a result, a convolutional neural network may learn a classification task automatically from photos, videos, texts, or audio, without the need for the manual feature extraction phase that humans typically perform.

A convolutional neural network can include hundreds or even thousands of hidden layers, each of which can learn different visual attributes, based on the multi-layer perceptron. The incoming data is filtered using several filters, and the convoluted output of one layer is sent into the next layer. Filters can start out simple (e.g., brightness, edges, etc.) and gradually get more complex, characterizing an item in a unique way. Convolution, ReLU, and pooling layers are the most commonly used layers in convolutional neural networks to learn these specific features. Finally, a classification phase follows this feature learning architecture, with predictions made using a dense layer in conjunction with a softmax layer.

$$\text{Softmax}\left(x_i\right) = \frac{\exp\left(x_i\right)}{\sum_j \exp\left(x_j\right)} \tag{2.10}$$

An overview for the convolutional neural network architecture is shown in Fig. 2.6, while the various mentioned layers are briefly explained in the following list:

- ***Convolution***: this layer applies a sequence of convolutional filters on the input image in order to learn certain features. Using the previously mentioned filters, a feature map is created during the convolution step, which can be thought of as a moving window applied to the entire input image. To generate the map inside the window, a dot product is applied to the pixels. In mathematicians language this operation is called "discrete multidimensional convolution"

$$\sum_{k_1=-\infty}^{\infty} \sum_{k_2=-\infty}^{\infty} \cdots \sum_{k_M=-\infty}^{\infty} h\left(k_1, \ldots, k_M\right) x\left(n_1 - k_1, \ldots, n_M - k_M\right) = x ** h \tag{2.11}$$

  where $**$ is the multi-dimensional convolution operator and $M$ is the number of dimensions, typically 2 or 3 [20] in case of grayscale or RGB images.

- ***ReLU***: by mapping negative values to zero while keeping positive values, this layer allows for a faster and more effective learning phase. After having linear operations in the previous levels, the rationale is to introduce some type of non-linearity.

$$f(x) = x^+ = \max(0, x) \tag{2.12}$$

- ***Pooling***: this layer is used in order to simplify the output by applying a non-linear subsampling and, consequently, reducing the number of parameters the network has to learn. This reduction can be obtained by using a filter on a feature map. Examples of filters for the pooling layer are: calculating the sum, using the maximum value, or computing the average, of the values found inside the filter window. An example pf pooling is given in Fig. 2.5.

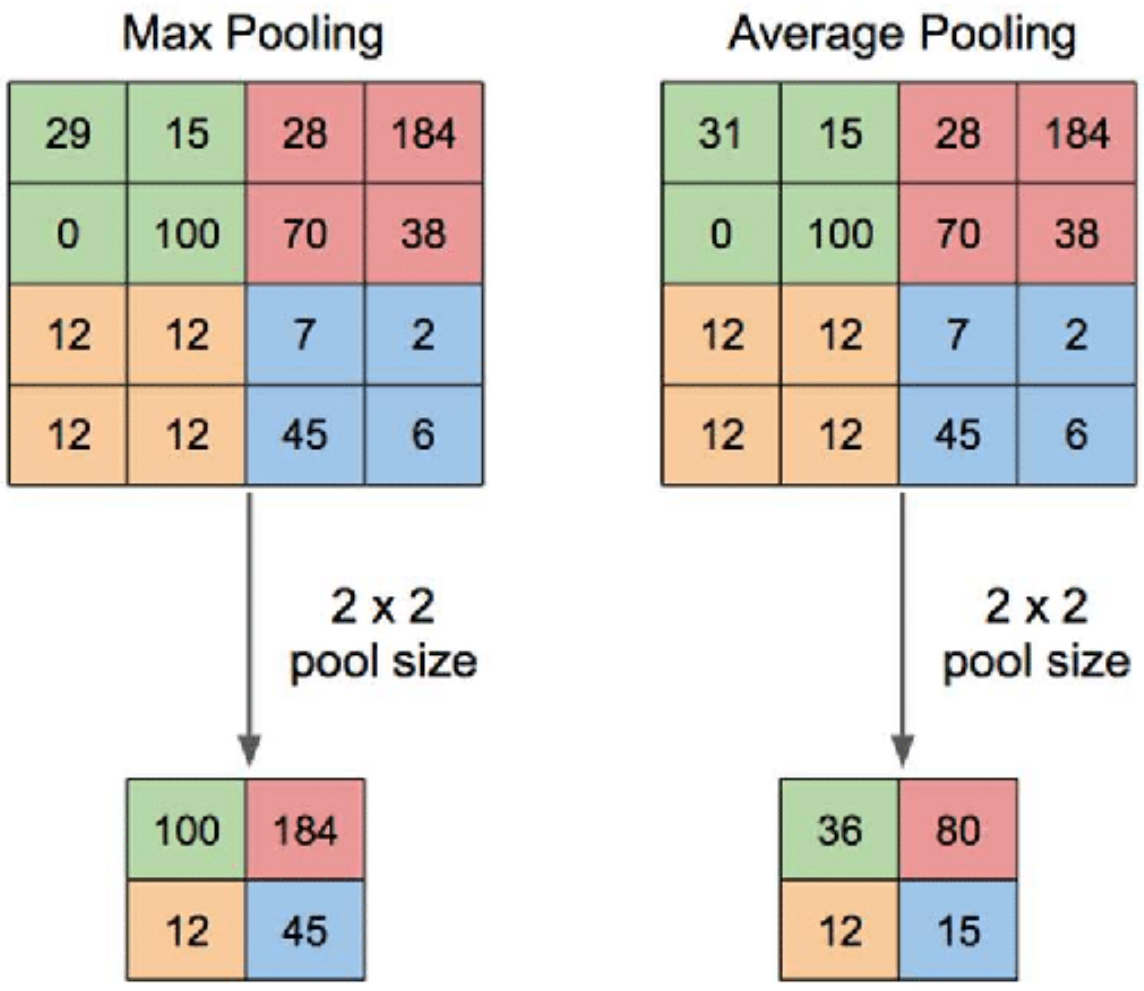

**Figure 2.5.** Examples of max-pooling and average-pooling.[2]

## 2.3 Efficient approaches

Computer vision, natural language processing, speech recognition, information retrieval and many other fields have all been transformed by deep learning. However, as deep learning models have improved, the number of parameters, latency, and resources required to train them have all increased dramatically. As a result, it's become critical to pay attention to a model's footprint measurements as well as its quality.

[2]Image taken from the site: https://www.researchgate.net/figure/Illustration-of-Max-Pooling-and-Average-Pooling-Figure-2-above-shows-an-example-of-max_fig2_333593451.

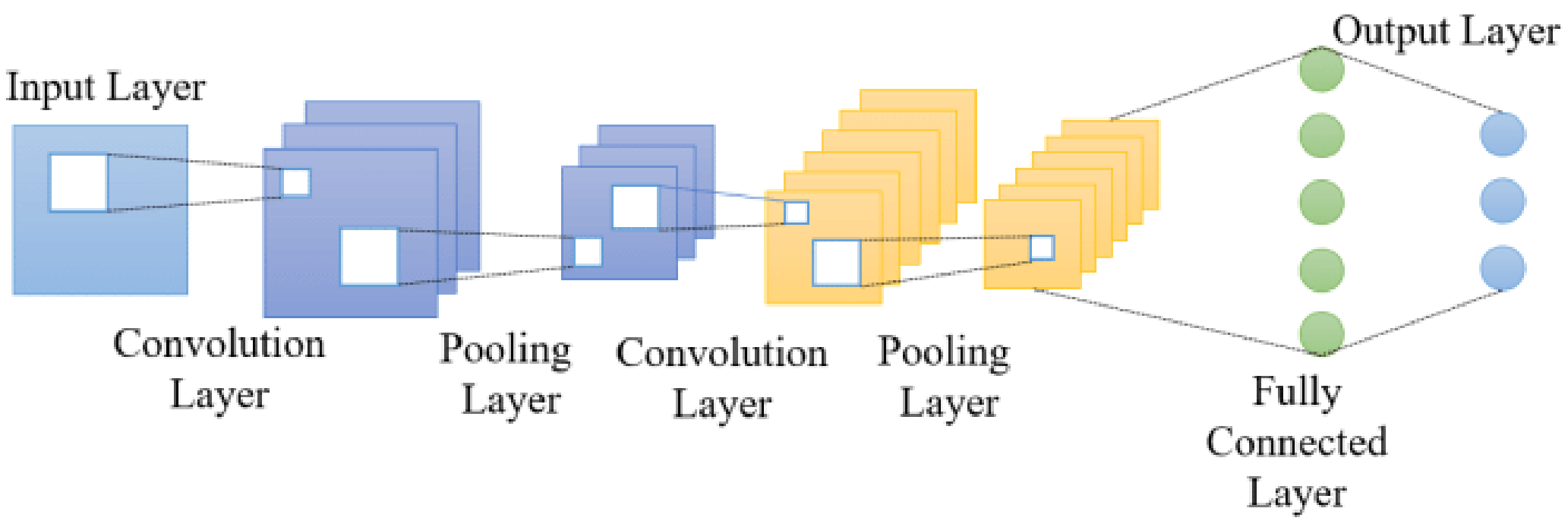

**Figure 2.6.** Convolutional neural network architecture example showing all of the described layers, namely the convolution, ReLU, pooling, dense, and softmax ones.[3]

The field of efficient deep learning concentrate on:

- ***Inference efficiency:*** addresses questions that someone who is deploying a model for inference could have, such as: is the model small? Is it fast enough? How many parameters do the model needs?

- ***Training efficiency:*** includes queries like: how long does it take to train a model? How many devices can the model use? Can the model be stored in memory? It could also include queries like how much data the model would require to obtain the desired results on the task.

In order to sort what literature made for efficient deep learning it is possible to split the strategies on some categories, that will later be analyzed more in depth:

- ***Compression techniques:*** these are generic strategies and procedures for optimizing the architecture of a model, usually by compressing its layers. Quantization [21], for example, seeks to compress a layer's weight matrices by reducing precision (e.g., from 32-bit floating point values to 8-bit unsigned integers) with little quality loss.

[3]Image taken from the site: https://www.researchgate.net/figure/Basic-architecture-of-CNN_fig3_335086346

- ***Learning techniques:*** these are algorithms that concentrate on training the model in a unique way (to make fewer prediction errors, require less data, converge faster, etc.). By reducing the number of parameters, the enhanced quality can be exchanged for a smaller footprint/more efficient model. Distillation [22] is an example of a learning strategy that allows a smaller model to improve its accuracy by learning to resemble a larger model.

- ***Automation:*** these are tools for automating the improvement of a model's basic metrics. Hyper-parameter optimization [23] is an example of how optimizing hyper-parameters can help boost accuracy, which can then be swapped for a model with fewer parameters. Similarly, architecture search [24] comes into this category, where the architecture is modified and the search aids in the discovery of a model that maximizes both loss/accuracy and another parameter such as model latency, model size, and so on.

- ***Efficient architectures:*** these are key building pieces that were created from the ground up (convolutional layers, attention, etc.) and represent a major improvement over the previous methodologies (fully connected layers, and recurrent neural networks respectively).

- ***Infrastructure:*** this includes model training frameworks like Tensorflow [25], PyTorch [26], and others (as well as tools for deploying efficient models like Tensorflow Lite (TFLite), PyTorch Mobile, and others). We rely on infrastructure and tooling to maximize the benefits of efficient models. For example, with quantized models, we need the inference platform to support common neural network layers in quantized mode to obtain both size and latency advantages.

### 2.3.1 Compression techniques

As previously stated, compression approaches are typically generic techniques for achieving a more efficient representation of one or more layers in a neural network, with a potential quality trade-off. In return for as little quality loss as possible, the efficiency goal could be to optimize the model for one or more of the footprint measures, such as model size, inference latency, training time necessary for convergence, and so on. These strategies can help increase model generalization in some circumstances where the model is over-parameterized.

- ***Pruning:*** pruning a neural network can be done in a variety of ways. Weights can be pruned. Setting individual parameters to zero and making the network sparse accomplishes this. The number of parameters in the model would be reduced while the architecture remained unchanged. You have the ability to delete entire nodes from the network. This would reduce the size of the network architecture while maintaining the accuracy of the larger network.

  Determining what to prune is a big difficulty in pruning. You want the parameters you remove from a model to be less useful if you're deleting weights or nodes. Different heuristics and strategies exist for evaluating which nodes are less significant and can be eliminated with little impact on accuracy. To evaluate how essential a neuron is for the model's performance, it is possible to utilize heuristics based on its weights or activations; the idea is to eliminate as many of the less important factors as possible.

  One of the most important factors to consider when pruning is when it should be placed in the machine learning training/testing cycle. However, you may notice that the model's performance has degraded as a result of the trimming. Fine-tuning, or retraining the model after pruning to regain accuracy, can fix this.

- ***Quantization:*** a typical network's weights and activations are almost entirely in 32-bit floating-point data. Reduce the accuracy of the weights and activations by quantizing to a lower-precision datatype (typically 8-bit fixed-point integers) is one way to reduce model footprint. We can benefit from quantization in two ways: (a) reduced model size and (b) reduced inference latency. When only the model size is a limitation, we can use a technique called weight quantization to enhance model size [27], as long as only the model weights are lowered in precision. To enhance latency, the activations must also be in fixed-point (activation quantization [21, 28], which means that all operations in the quantized graph must be done in fixed-point math.

- ***Other compression techniques:*** other compression approaches, such as low-rank matrix factorization [29], K-means clustering [30], weight-sharing [31], and others, are also in use for model compression and may be ideal for compressing hotspots in a model further.

### 2.3.2 Learning techniques

Learning techniques attempt to train a model in a different way in order to achieve higher quality measures while complementing, or in some cases replacing, standard supervised learning. By lowering the number of parameters and layers in the model and achieving the same baseline quality with a smaller model, the improvement in quality can sometimes be traded off for a smaller footprint. One advantage of paying attention to learning techniques is that they are only used during training and have no bearing on inference.

- ***Distillation:*** the use of ensembles to aid generalization is widely established [3, 32]. The idea is that this allows for the learning of numerous independent hypotheses, which is thought to be more efficient than learning a single hypothesis. [33] discusses standard ensemble methods like bagging (learning models that are trained on non-overlapping data and then ensembling them), boosting (learning models that are trained to fix the classification errors of other models in the ensemble), averaging (voting by all ensemble models). [34] employed huge ensembles to label synthetic data created by a variety of methods. The synthetic data is then fed into a smaller neural network that is trained to learn not only from the labeled data but also from the weakly labeled data. Single neural nets were shown to be capable of simulating the performance of bigger ensembles while being 1000 times smaller and faster. This demonstrated that the cumulative knowledge of ensembles may be transferred to a single small model. However, relying solely on existing labeled data may not be sufficient. It is conceivable to adapt the distillation concept to deal with teachers' and students' intermediate outputs. Between the instructor and student convolutional networks, [35] transfer intermediate 'attention mappings'. The idea is to get the learner to concentrate on the sections of the image that the teacher is looking at. [36] employs a progressive-knowledge transfer technique in which they distill the BERT student and instructor models layer by layer, but in phases, with the first l layers distilled at the l-th stage. They get a x4.3 smaller and x5.5 quicker BERT with little quality losses by combining previous design advancements.

- ***Data augmentation:*** the size of the training data corpus correlates with improved generalization when training big models for complicated tasks in a Supervised Learning regime. [143] [37] shows a logarithmic improvement in prediction accuracy as the number of labeled cases increases. However, acquiring high-quality labeled data frequently necessitates the involvement of a human and might be costly. Data augmentation is a clever approach of dealing with the paucity of labeled data by artificially increasing the existing dataset using augmentation methods. These augmentation methods are modifications that can be applied cheaply to the given instances, resulting in the new label of the enhanced example not changing or being inferred quickly. Transforming an image of a dog horizontally or vertically by a minor number of pixels, rotating it by a small angle, and so on would not materially modify the image, hence the classifier should still categorize the transformed image as 'dog.' This forces the classifier to acquire a more robust image representation that is more generalizable across various modifications. The changes mentioned here have been shown to increase the accuracy of convolutional networks for a long time [2,38].

- ***Self-supervised learning:*** labeled data is frequently used in the supervised-learning paradigm. As previously said, it necessitates human interaction and is also costly. The amount of labeled data required to attain good quality on a non-trivial task is also significant. While approaches like as Data-augmentation, distillation, and others can help, they all require the presence of labeled data to reach a baseline performance. By attempting to extract additional supervisory bits from each example, self-supervised learning reduces the necessity for labeled data in order to learn generalized representations. It does not need to focus solely on the label because it focuses on learning robust representations of the example itself. Because unlabeled data is abundant in many domains (e.g., books, Wikipedia, and other text for natural language processing, web images and videos for computer vision, and so on), the model would not be slowed down by it when learning to tackle various pretext tasks. After learning general representations that transfer well across tasks, the models can be fine-tuned with labeled data and customized to handle the target task by adding certain layers that project the representation to the label space. Because the labeled data will be used to learn how to translate the high-level representations into the label space rather than learning elementary features, the amount of labeled data will be a fraction of what would have been required to train the model from scratch. Fine-tuning models pre-trained with self-supervised learning are data-efficient (they converge faster, achieve greater quality for the same amount of labeled data, etc.) from this perspective [2].

### 2.3.3 Automation

It is possible to delegate some efficiency-related labor to automation, with automated systems searching for ways to train more efficient models. Apart from lowering human workload, it also reduces the bias that manual decisions may create in model training, as well as searching for best solutions systematically and automatically. The trade-off is that these approaches may take a lot of computing power, therefore they must be used with caution.

- ***Hyper-parameter optimization:*** [39] is one of the most often utilized strategies in this area. For faster convergence, hyper-parameters like initial learning rate, weight decay, and so on must be fine-tuned. They can also choose the network design, such as how many fully connected layers there are, how many filters there are in a convolutional layer, and so forth. Experimentation can help us get a sense of the parameter's range, but finding the best values necessitates a search for the exact values that optimize the specified objective function (typically the loss value on the validation set). With the growing number of hyper-parameters and/or their conceivable values, manually searching for them gets tiresome. As a result, let us investigate possible search-automation algorithms. Grid search (also known as parameter sweep) is a simple algorithm for automating hyper-parameter optimization. Because each trial is independent of the others, each trial can be run in parallel, and the optimal combination of hyper-parameters can be discovered once all of the trials have been finished. Because this method tries all conceivable combinations, it suffers from the dimensionality curse, which causes the total number of trials to rapidly rise. Random search [40] is another strategy, in which trials are picked at random from the search space. Each trial can still be run at random because it is independent of the others. The search is led by actively estimating the value of the objective function at different places in the search space, and then spawning trials based on the information obtained so far in bayesian optimization based search [41]. The goal function is estimated using a surrogate function that begins with a prior estimate.

- ***Neural architecture search:*** it is possible to think of neural architecture search as an extension of hyper-parameter optimization in which we look for parameters that affect the network architecture. The literature [42] agrees on categorizing neural architecture search as a system made up of the following components: search space, search algorithm and evolution strategy. The user is expected to encode the search space either explicitly or indirectly. We might think of this as a 'controller' that generates sample candidate networks in conjunction with the search method. Following that, these applicants will be trained and evaluated for fitness at the evaluation stage. This fitness value is then fed back to the search algorithm, which uses it to improve candidate generation. While the execution of each of these pieces varies, the overall framework is consistent throughout the seminal work in this field. [24] revealed that reinforcement learning may be used to create end-to-end neural network topologies. The controller in this case is a recurrent neural network, which creates the architectural hyper-parameters of a feed-forward network one layer at a time, such as the number of filters, stride, and filter size, among other things. In Fig. 2.7 it is possible to see a general procedure for neural architecture search.

[4] Image taken from the site: https://www.researchgate.net/figure/General-procedure-of-neural-architecture-search_fig1_353166978

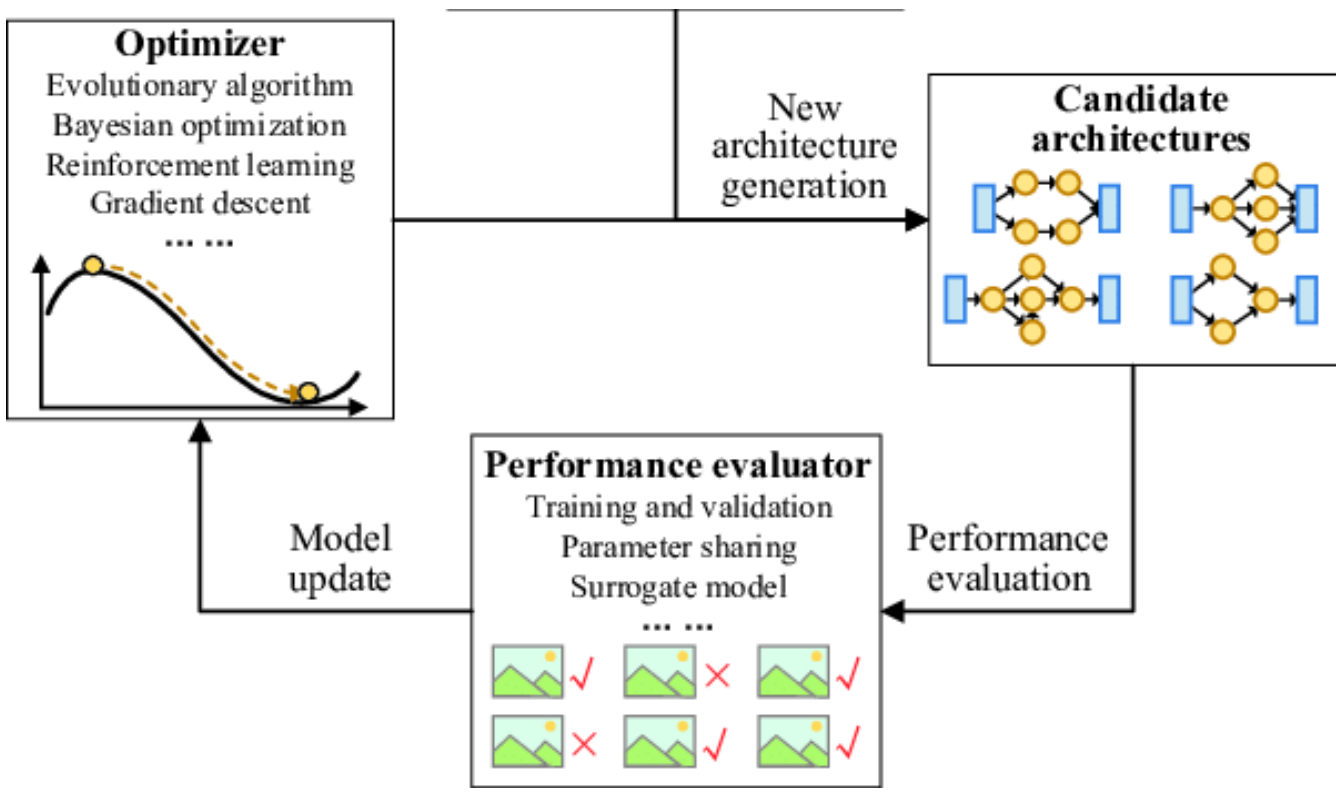

**Figure 2.7.** Neural architecture search general procedure.[4]

# Chapter 3

# Optimization

This chapter is organized as follows. In section 3.1 the evolution and the theory under the gradient descent algorithms are presented. In section 3.2 the strategy to train neural networks with genetic and evolutionary algorithm are explored.

## 3.1 Gradient descent

Gradient descent is one of the most widely used optimization algorithms, and it is by far the most frequent method for optimizing neural networks. At the same time, every current deep learning library includes implementations of multiple gradient descent optimization methods. However, because practical descriptions of their merits and weaknesses are difficult to come by, these algorithms are frequently used as black-box optimizers. Gradient descent is a way to minimize an objective function $J(\theta)$ parameterized by a model's parameters $\theta \in \mathbb{R}^d$ by updating the parameters in the opposite direction of the gradient of the objective function $\nabla_0 J(\theta)$ w.r.t. to the parameters. The learning rate $\eta$ determines the size of the steps we take to reach a (local) minimum. In other words, we follow the direction of the slope of the surface created by the objective function downhill until we reach a valley.

### 3.1.1 Batch gradient descent

The gradient of the cost function w.r.t. the parameters $\theta$ for the full training dataset is computed using vanilla gradient descent, also known as batch gradient descent:

$$\theta = \theta - \eta \cdot \nabla_\theta J(\theta; x; y) \tag{3.1}$$

where $x$ are the training samples and $y$ are the labels.

Batch gradient descent is difficult and intractable for datasets that do not fit in memory since we need to calculate the gradients for the entire dataset to execute just one update. Batch gradient descent also prevents us from updating our model in online mode. It is guaranteed to converge to the global minimum for convex error surfaces and to a local minimum for non-convex surfaces.

### 3.1.2 Stochastic gradient descent

In contrast, stochastic gradient descent updates a parameter for each training example $x^{(i)}$ and label $y^{(i)}$:

$$\theta = \theta - \eta \cdot \nabla_\theta J\left(\theta; x^{(i)}; y^{(i)}\right) \tag{3.2}$$

For big datasets, batch gradient descent performs redundant calculations by recalculating gradients for similar cases before each parameter update. By executing one update at a time, stochastic gradient descent eliminates redundancy. As a result, it is usually significantly faster and can also be utilized for online learning. stochastic gradient descent makes frequent, high-variance updates, causing the objective function to vary a lot as can be seen in Fig. 3.1.

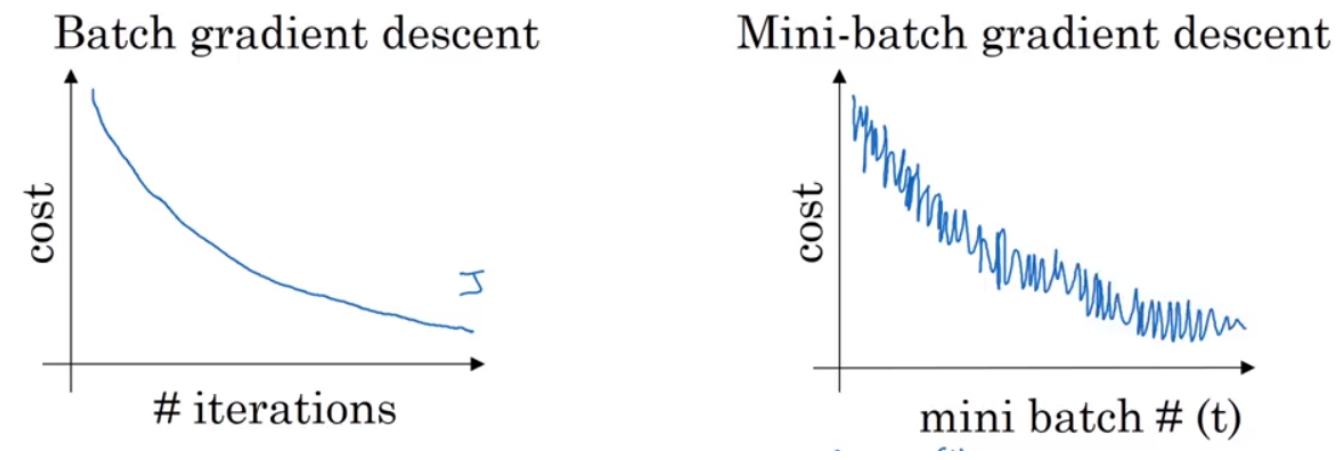

**Figure 3.1.** Stochastic gradient descent fluctuation vs batch gradient descent fluctuation.[1]

### 3.1.3 Mini-batch gradient descent

Finally, mini-batch gradient descent combines the best of both worlds by updating each mini-batch of $n$ training examples:

$$\theta = \theta - \eta \cdot \nabla_\theta J\left(\theta; x^{(i:i+n)}; y^{(i:i+n)}\right) \tag{3.3}$$

This way, it decreases the variance of parameter updates, which can lead to more steady convergence; and can take advantage of highly optimized matrix optimizations seen in modern deep learning libraries, which make computing the gradient w.r.t. a mini-batch quite efficient. Mini-batch sizes typically range from 50 to 256, but this might vary depending on the application. When training a neural network, mini-batch gradient descent is frequently the algorithm of choice, and the name stochastic gradient descent is commonly used when mini-batches are used.

### 3.1.4 Mini-batch gradient descent momentum

Stochastic gradient descent has problems navigating saddle points, which are widespread around local optima and are regions where the landscape energy slopes considerably more steeply in one dimension than in another [43]. As shown in Fig. 3.2, stochastic gradient descent oscillates over the saddle point's slopes while only making slow progress near the bottom toward the local

[1]Image taken from the site: https://upload.wikimedia.org/wikipedia/commons/f/f3/Stogra.png

optimum.

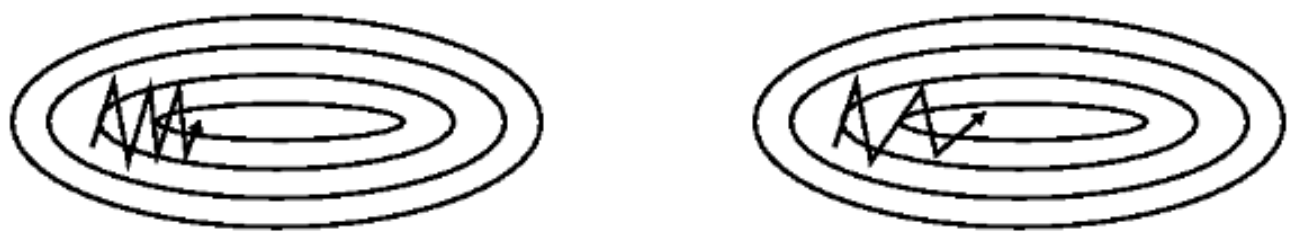

**Figure 3.2.** Left stochastic gradient descent without momentum, right stochastic gradient descent with momentum.[2]

.

### 3.1.5 Adam

Adaptive Moment Estimation (Adam) [44] is a method that computes adaptive learning rates for each parameter. Adam keeps an exponentially decaying average of past gradients

$$\begin{aligned} m_t &= \beta_1 m_{t-1} + (1 - \beta_1)\, g_t \\ v_t &= \beta_2 v_{t-1} + (1 - \beta_2)\, g_t^2 \end{aligned} \tag{3.4}$$

where $m$ and $v$ are moving averages, $g$ is the gradient on current mini-batch and $\beta_1 \beta_2$ are the parameters introduced with Adam. $m_t$ and $v_t$ are initialized as a vector with all 0. The authors of Adam see that they are biased towards zero, so they compute bias-corrected first and second moment estimations to compensate:

$$\begin{aligned} \hat{m}_t &= \frac{m_t}{1 - \beta_1^t} \\ \hat{v}_t &= \frac{v_t}{1 - \beta_2^t} \end{aligned} \tag{3.5}$$

They then use these to update the parameters

$$\theta_{t+1} = \theta_t - \frac{\eta}{\sqrt{\hat{v}_t} + \epsilon} \hat{m}_t \tag{3.6}$$

[2]Image taken from the site: https://www.willamette.edu/~gorr/classes/cs449/momrate.html

## 3.2 Training through evolutionary algorithms

Evolutionary algorithms [45], are a type of stochastic optimisation bio-inspired algorithm that uses evolutionary principles to build long-lasting adaptive systems. Three pioneering evolutionary approaches are genetic algorithms [46], evolution strategies [47] and evolutionary programming [48]. The versatility of these algorithms is a key characteristic, since it allows to blend elements from two or more evolutionary algorithm techniques.

Evolutionary algorithms work with a group of potential solutions to a certain problem. Each potential solution represents a place in the search space where the best solution can be found. Over a number of generations, the population is evolved using genetic operators to produce superior results to the challenge.

A fitness function is applied to each member of the population to determine how excellent or awful the proposed solution is for the problem at hand. The fitness value assigned to each individual in the population determines how successful the individual will be in passing on (part of) their code to future generations. Higher values (for maximisation problems) or lower values will be allocated to better performing solutions (for minimisation problems). Genetic operators are used to carry out the evolutionary process. Most evolutionary algorithms have operators that select individuals for reproduction, produce new individuals depending on the selected individuals, and eventually determine the composition of the population in the next generation. Most evolutionary algorithms paradigms use selection, crossover, and mutation as important genetic operators. The selection operator is in charge of selecting one or more people from the population based on their fitness levels. There have been multiple selection operators proposed. Tournament selection is one of the most common selection operators due of its simplicity. The goal is to choose the best individual from a population pool. The stochastic crossover operator, also known as recombination, swaps material between two individuals. This operator is in charge of making the most of the available search space. The stochastic mutation operator causes random alterations to an individual's genes. The exploration of the search space is the responsibility of this operator. The mutation operator is critical for maintaining population variety and retrieving genetic material lost throughout evolution.

The previously described evolutionary process is repeated until a requirement is met. Typically, until a certain number of generations have been completed. The population of the latest generation is the consequence of several generations of exploring and using the search space. It may also represent the global optimum solution because it encompasses the best evolved potential solutions to the problem.

### 3.2.1 Genetic algorithms

[46] introduced this evolutionary algorithm in the 1970s. This was owing to exceptional outcomes as well as reaching out to a wide range of academic communities, including machine learning and neural networks. Although genetic algorithms were once thought of as function optimizers, they are now more commonly thought of as search algorithms capable of finding near-optimal solutions. In the specialized literature, various types of genetic algorithms have been proposed. One of the most common encodings used in genetic algorithms is the bitstring fixed-length representation very similar to DNA concept. Crossover, as the primary genetic operator, and mutation, as the secondary genetic operator, create offspring during the evolutionary process. In fig. 3.3 it is possible to see a flow chart representing the phases of a genetic algorithm.

### 3.2.2 Evolution strategies

[47] introduced these evolution algorithms in the 1960s. Evolution strategies are frequently used to model real-valued optimisation problems. In evolution

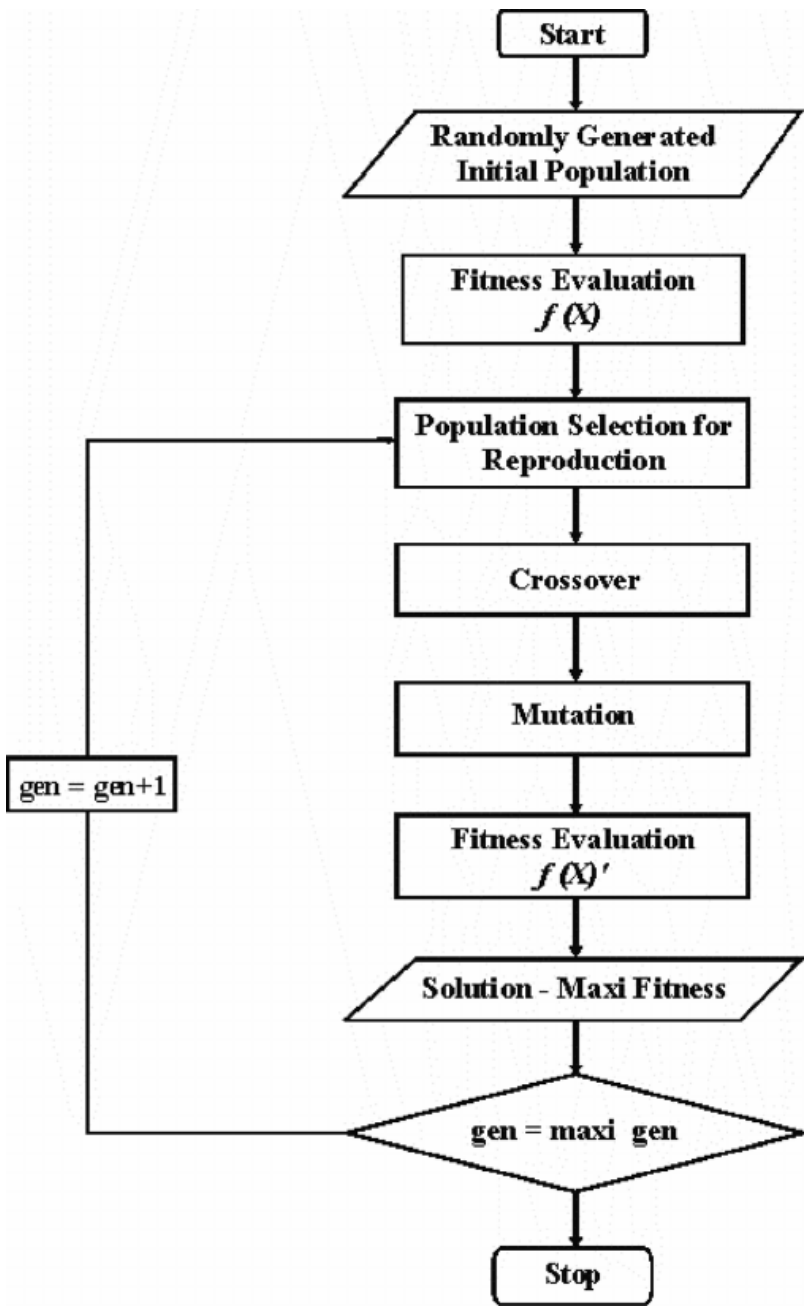

**Figure 3.3.** General procedure of a genetic algorithm.[3]

strategies, the main operator is mutation, whereas the secondary, optional operator is crossover. The ($\mu$, $\lambda$)-ES refers to the algorithms where $\mu$ refers to the size of the parent population, whereas $\lambda$ refers to the number of offspring that are produced in the following generation before selection is applied. Nowadays, Hansen's covariance matrix adaptation-evolution strategies [49–51], which adapts the whole covariance matrix of a normal search (mutation) distribution, is the state of the art evolutionary strategies.

### 3.2.3 Evolutionary programming

[48] proposed these evolution strategies in the 1960s, and there are relatively few differences between evolutionary strategy and evolutionary programming. The primary difference between these two evolution algorithms paradigms is that evolutionary programming does not use crossover, whereas evolutionary strategy uses it as a secondary and rarely used genetic operator. Another

[3] Image taken from the site: https://www.researchgate.net/figure/General-procedure-of-genetic-algorithm_fig4_233754686

contrast is that in evolutionary programming, $M$ parents usually generate $M$ children, whereas in evolutionary strategy, genetic operators create more offspring than their parents.

Over the last few decades, backpropagation has been one of the most successful and dominating strategies for training neural networks. Stochastic gradient descent is applied to the weights of the neural networks in this simple, effective, and elegant method, with the goal of keeping the overall error as low as possible. However, [52] pointed out, the widely accepted view, was that backpropagation would lose its gradient within deep neural networks. This was later proven to be incorrect, because backpropagation and stochastic gradient descent are effective in optimizing deep neural networks even when there are millions of connections. Backpropagation and stochastic gradient descent both benefit from the availability of enough training data as well as computational capability. The success of stochastic gradient descent in deep neural networks is still remarkable in a problem space with so many dimensions. stochastic gradient descent should be particularly vulnerable to local optima in practice [52].

# Chapter 4

# Proposed approach

This chapter is organized as follows. Section 4.1 shows the related works and the attempts to train a variant full binary neural network with genetic and evolutionary algorithms made by the candidate. In section 4.2 the final binary neural network's architecture and training strategy are explained.

## 4.1 Genetic optimizer

The candidate has experimented different black-box optimization genetic algorithms instead of gradient descent; although the network differs slightly from a classic binary neural network, the goal of this section is to find a reliable gradient descent replacement in a full binary network.

### 4.1.1 Training strategy

This is part of the related work, and one of the goals of the thesis is to determine whether there is a fully binary algorithm capable of training a neural network with a usable convergence speed.

The first kind of investigation is about the literature about the Signum algorithms [53]: this optimizer uses the function *sign* immediately before updating the weights, as can be seen in Fig. 4.1, but performs all previous

**Algorithm** SIGNUM

**Input:** learning rate $\delta$, momentum constant $\beta \in (0,1)$,
current point $x_k$, current momentum $m_k$

$\tilde{g}_k \leftarrow \text{stochasticGradient}(x_k)$

$m_{k+1} \leftarrow \beta m_k + (1-\beta)\tilde{g}_k$

$x_{k+1} \leftarrow x_k - \delta \, \text{sign}(m_{k+1})$

**Figure 4.1.** Algorithm using *sign* function before weights update [53]

calculations with full precision. So, the true benefit of Signum is in distributed training, where communication between machines is critical.

None of the benefits of these approaches are useful for making training less computationally intensive, faster, or less memory greedy, so a different different approach can be investigated.

### 4.1.2 Architecture

The architecture is a neural network with 2 to 5 layer without batch normalization. The layer is a binary fully connected built with the XOR-bitCount operation. In figure 4.2 the architecture is shown, where $\oplus$ is the XOR oper-ator and majority voting returns 0 if the majority of the inputs are 0 and 1 otherwise.

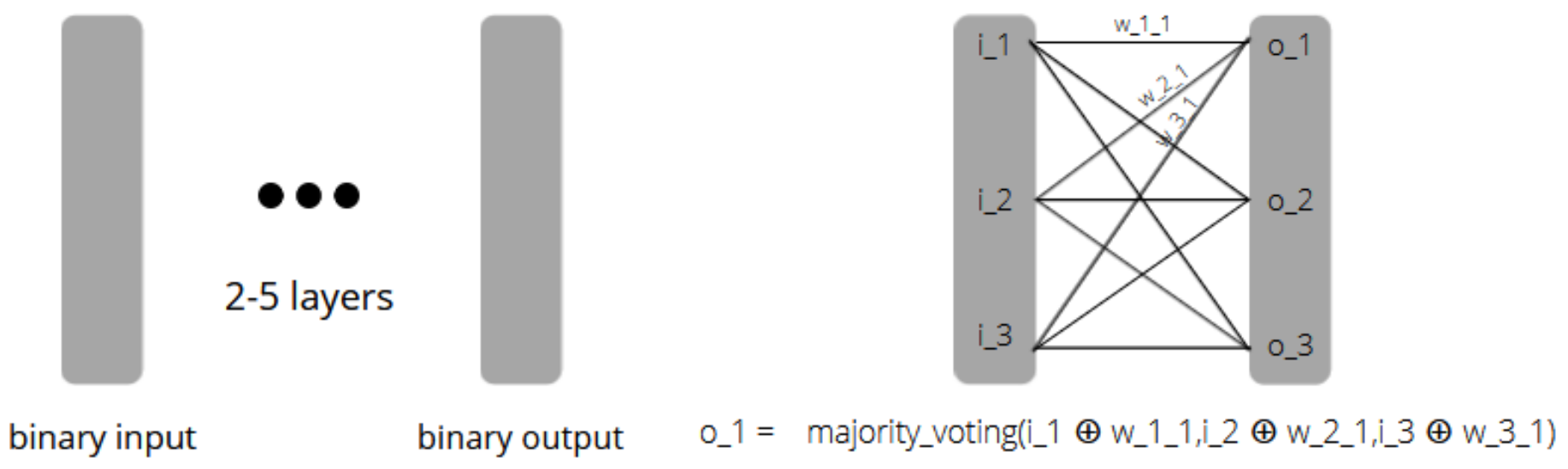

**Figure 4.2.** Network's architecture for the training based on genetic algorithm

### 4.1.3 Dataset

MNIST (5.1.1) is the dataset used for the tests. Because of the nature of the data, this dataset was chosen. After applying the function *sign*, the images in the dataset lose very little information: as can be seen in Fig. 4.3, almost all the pixels are white or black.

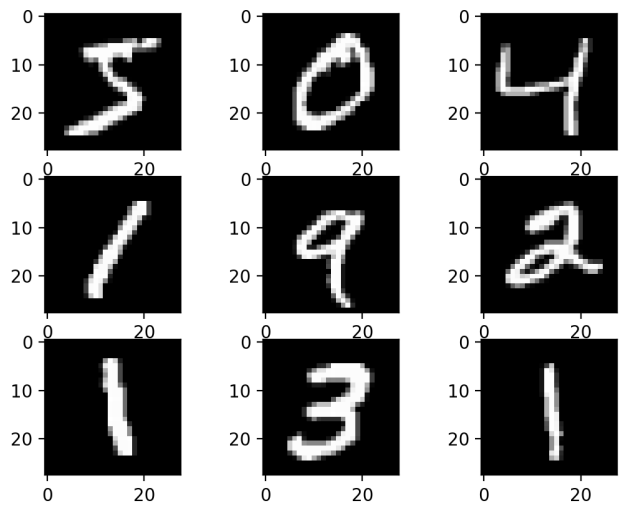

**Figure 4.3.** MNIST samples overview.

### 4.1.4 Majority loss

Due to the binary output restriction, the final loss is composed of groups of $k$ bits representing labels, where the number of 1 in each group represents the sample's membership to a label. As you can see in 4.4, the predicted class is the one with the most 1s.

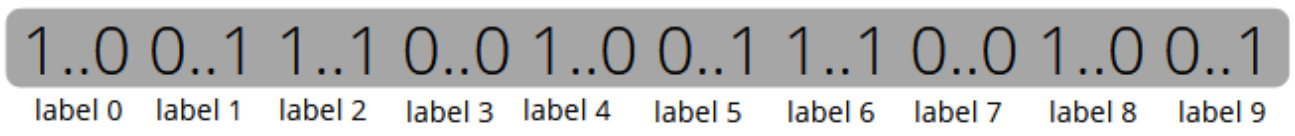

**Figure 4.4.** An example of binary loss.

When the count of one in two groups is equal, how should the predicted class be chosen? To make the problem smaller, increasing the number of bits in each group lowers the probability of getting this problem.

### 4.1.5 Algorithms

The algorithms seen are three:

- ***Naive perturbation evolutionary .1***: this algorithm is inspired by neural network adversarial attack by noise [54]. The concept under this algorithm is random flip weights with a certain probability and check if the model is more accurate.

- ***Evolutionary algorithm .2***: this a candidate's reformulation of [55]. The concept under this algorithm is random flip weights with a certain probability starting from elite parents and the output of one step is a new elite. The elite are selected with a score that is a linear combination of current accuracy on test of the model and accuracy of the model's ancestors. The difference with [55] is the application: in this work is applied to a deeper and larger network.

- ***Counting error evolutionary algorithm' .3***: this is a candidate's original algorithm. The concept under this algorithm is the counting of the errors of the outputs in the batch; a bit in the the output is *wrong* if the bit is at least $BatchSize/2$ times not equal to the label in the batch. A bit in input of the last layer is set *wrong* if in the various majority voting contributed at least $BatchSize/2$ times to predict the a bit set *wrong*. The same process is continued until the input layer is reached. Finally the weights set *wrong* random flip with a certain probability. The idea of this algorithm is to emulate the idea of backpropagation but without derivative and floating point.

The algorithms experimented can be found on appendix 6.

### 4.1.6 Results

These related works contribute to a better understanding of the role of the optimizer in a very constrained setup; after 30 minutes, the execution of all the experiments are terminated. None of the tests performed with these algorithms or by hand on the hyper-parameters yielded an acceptable top-1 accuracy on MNIST, considering that using support vector machine [56] the accuracy is near 90% after some seconds [57].

With algorithm .1, the validation top-1 accuracy never got higher than random choice, 10%, and the convergence speed is slow.

With algorithm .2, the validation top-1 accuracy never got higher than 48%, and the convergence speed is slow.

The configuration with 5 layers, 100 bits for each label, and algorithm .3 had the highest accuracy, with a 60% top-1 accuracy. However, no further improvements were found after a thorough manual search.

## 4.2 Proposed binary neural network

The proposed network is a compound of tricks and improvements that this section will explain, the final architecture is shown in Fig. 4.11.

### 4.2.1 Baseline

The proposed network is based on ResNet-18 [58], with the architecture shown in Fig. 4.5. The success of residual networks is due to the inclusion of the so-called residual (often named also skip or shortcut) connections in the network. This allows the solution to be refined across the network's depth.

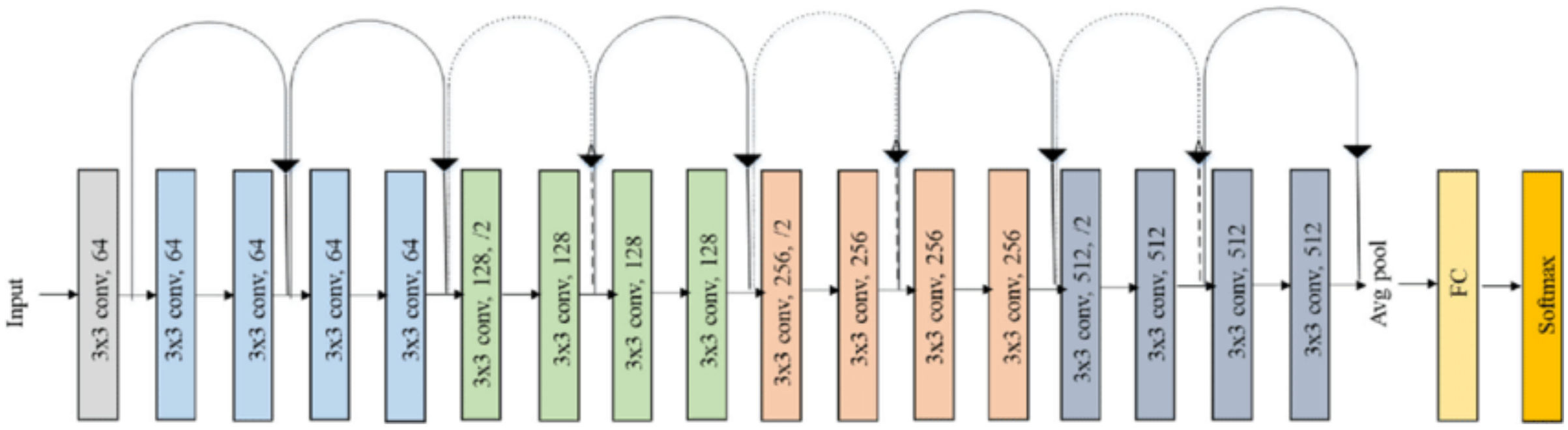

**Figure 4.5.** ResNet-18 architecture[1].

### 4.2.2 Preactivation

Preactivation [59], which changes the order of processes and achieves better precision and faster convergence, is another modification that advances the network's performances. The re-ordering of the operations, with respect to the original ResNet, is shown in Fig. 4.6

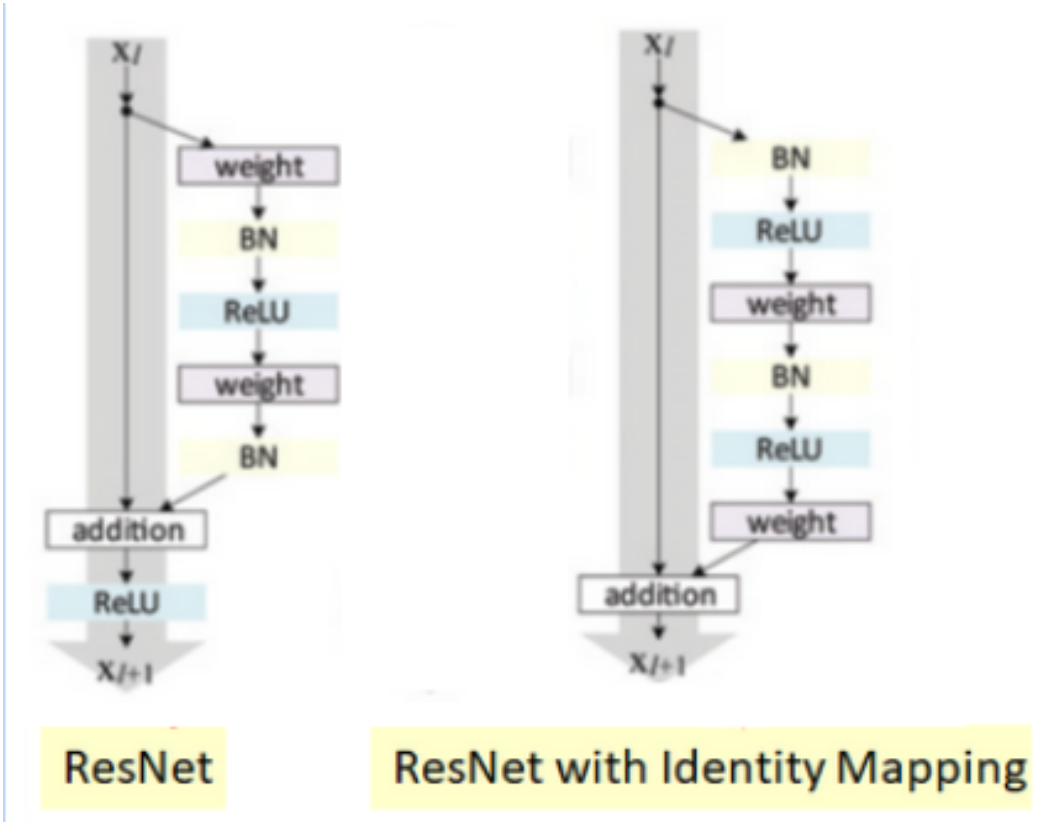

**Figure 4.6.** Pre-activation changes from Microsoft researches [59].

### 4.2.3 ReCU

The main issue with binary neural networks is the training: how can the networks be trained while simultaneously minimizing loss and quantization error? In probabilistic terms, BNN only knows whether or not to flip weights, hence flipping is the foundation of the training. Weights tend to flip too often or too infrequently when using Optimizers like Adam or SGDM. The trained weights are not further away from the initial value with SGDM; instead, with Adam, the weights might assume values that are much higher or lower, as shown in Fig. 4.7. Both are suffering from a lack of landscape energy exploration. The solution of this problem is ispired by ReCU's work [61].

$$\mathrm{ReCU}(w) = \max\left(\min\left(w, Q_{(\tau)}\right), Q_{(1-\tau)}\right) \tag{4.1}$$

[1]Image taken from the site: https://www.researchgate.net/figure/Original-ResNet-18-Architecture_fig1_336642248

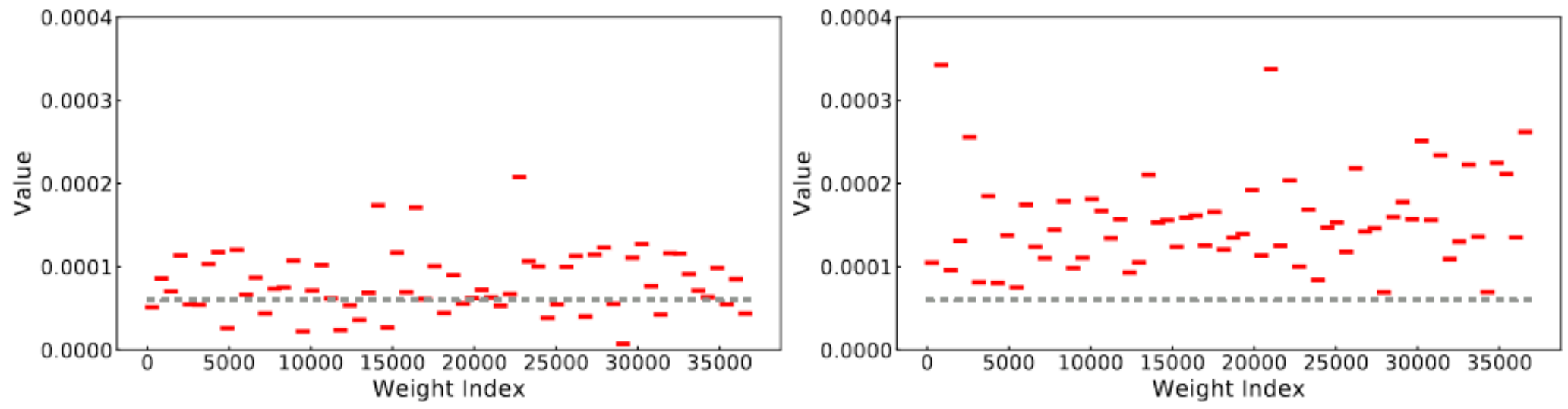

(a) Binary network with SGD (b) Binary network with Adam

**Figure 4.7.** Weights update in a convolutional neural network after one backward step with SGDM and Adam, from [60].

The weights are constrained to a range that changes throughout the training, that improves the precision thanks to the deeper exploration. They used this method with SGDM in their original work, however employing Adam in binary neural network's training is a beneficial practice due to the greater weights changes and quicker convergence [60].

### 4.2.4 Horizontal shift

Using horizontal shift of hardtanh allows the network to get higher accuracy [62]; according to them, binary neural networks often lack non-linearity in the model due to simple activation functions, but the extra shift of activation functions increases non-linearity.

### 4.2.5 Cyclic low precision training

The work's key contribution is the introduction of low bit training, which involves reducing the number of bits used for weights, activations, and gradients. In particular, the suggested solution employs a cyclic low-precision scheme inspired by work performed on multi-GPU systems [63]. In a cyclic mode, the objective is to lower the bits for weights and activations during the training. The proposed model, in particular, employs the 3-8 method, as seen in Fig. 4.8, that consists in cycles of 3 to 8 bits.

[1]Image taken from the site: https://www.researchgate.net/figure/Original-ResNet-18-Architecture$_{fig}1_{3}$36642248

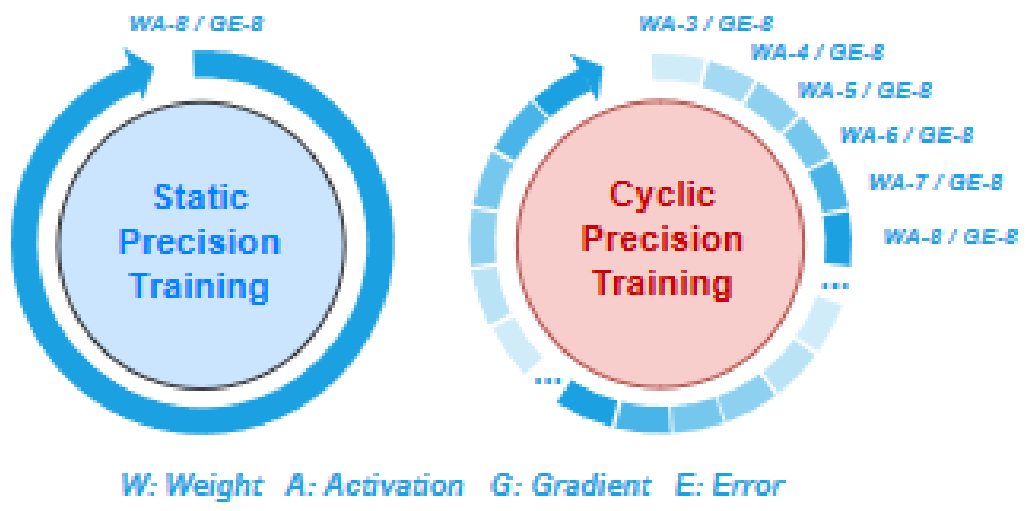

**Figure 4.8.** Left: static 8 bits training. Right: cyclic 3-8 bits training [63].

### 4.2.6 Cosine annealing learning rate scheduler

Cosine annealing is a type of learning rate schedule that has the effect of starting with a high learning rate and rapidly decreasing to a low value before rapidly increasing again. The resetting of the learning rate acts as a simulated restart of the learning process, and the use of good weights as the restart's starting point is referred to as a "warm restart", as opposed to a "cold restart", which may use a new set of small random numbers as a starting point. This scheduler make faster the convergence as can be seen in Fig. 4.9.

The following equation represents the function that regulates the learning rate, where $\eta^i_{\min}$ and $\eta^i_{\max}$ are ranges for the learning rate, and $T_{cur}$ account for how many epochs have been performed since the last restart:

$$\eta_t = \eta^i_{\min} + \frac{1}{2}\left(\eta^i_{\max} - \eta^i_{\min}\right)\left(1 + \cos\left(\frac{T_{cur}}{T_i}\pi\right)\right) \tag{4.2}$$

### 4.2.7 Data augmentation

The augmentation used in this proposed method is the same as ReCU [61], including random rotation, random horizontal flip and random crop with padding.

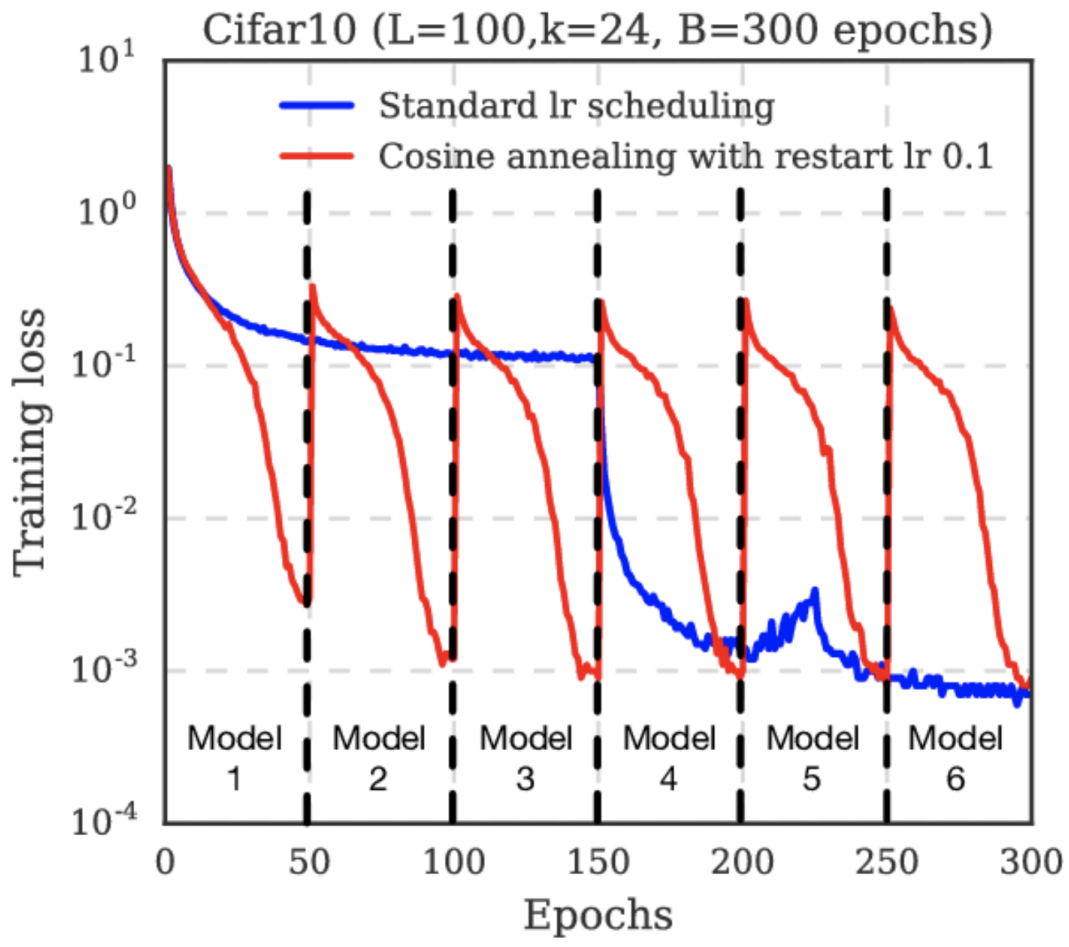

**Figure 4.9.** Fixed learning rate vs. cosine annealing[2].

### 4.2.8 Proposed architecture

Starting from the ResNet design, the first step to bring the network into the domain of binary inference is to make function activation binarizable and binarize the weight and the activation of each convolution. The literature

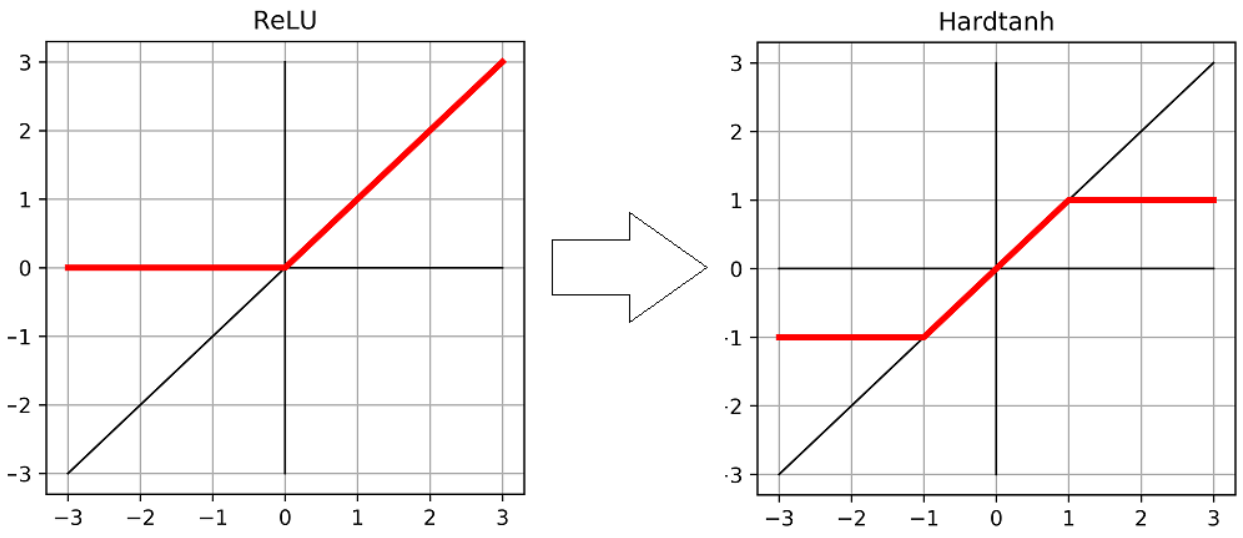

**Figure 4.10.** ReLU vs hardtanh.

shows how hardtanh reduce the quantization error [64]. An overview of the activation functions considered is shown in Fig. 4.10. Finally the proposed network is explained in Fig. 4.11

[2]Image taken from the site: https://paperswithcode.com/method/cosine-annealing

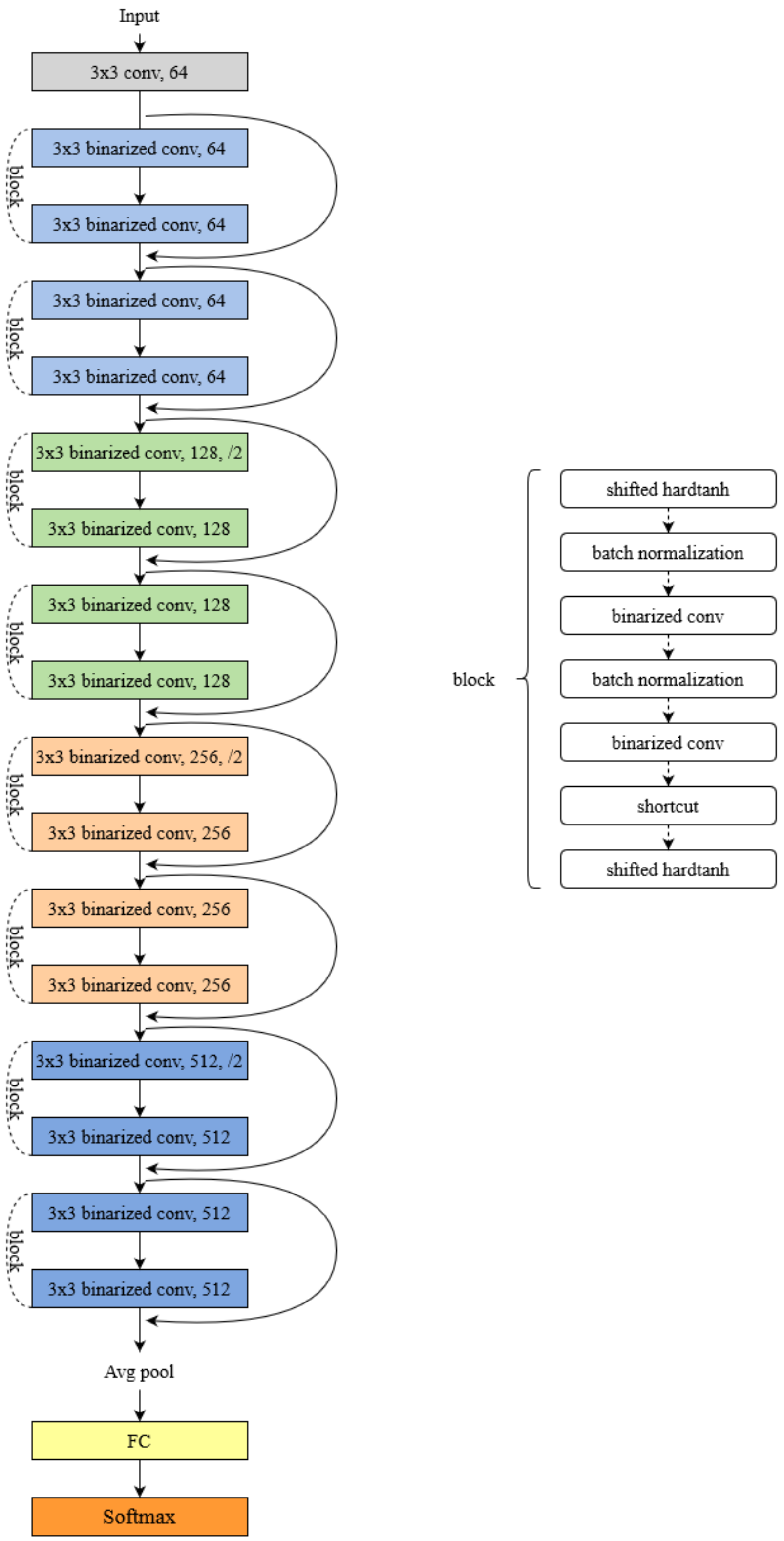

**Figure 4.11.** Overview of the proposed architecture. A block is composed by multiple layer as you can see in the right of the figure.

# Chapter 5

# Experiments

This chapter is organized as follows. In section 5.1 the datasets are presented. In section 5.2 tools and frameworks used are explained. In section 5.3 the results of the experiment are compared with the other state of the art techniques. Finally, in section 5.4, a recap of the presented work is provided and a comparison of performances during training is also provided.

## 5.1 Datasets

### 5.1.1 MNIST

A training set of 60,000 samples and a test set of 10,000 examples are available in the MNIST [65] database of handwritten digits. It's a subset of the NIST's wider collection. In a fixed-size image, the digits have been size-normalized and centered, as can be seen in Fig. 5.1.

It's a useful database for who wants to experiment with machine learning and pattern recognition techniques on real-world data with minimal preparation and formatting.

[0]Image taken from the site: https://it.wikipedia.org/wiki/MNIST$_{d}atabase$

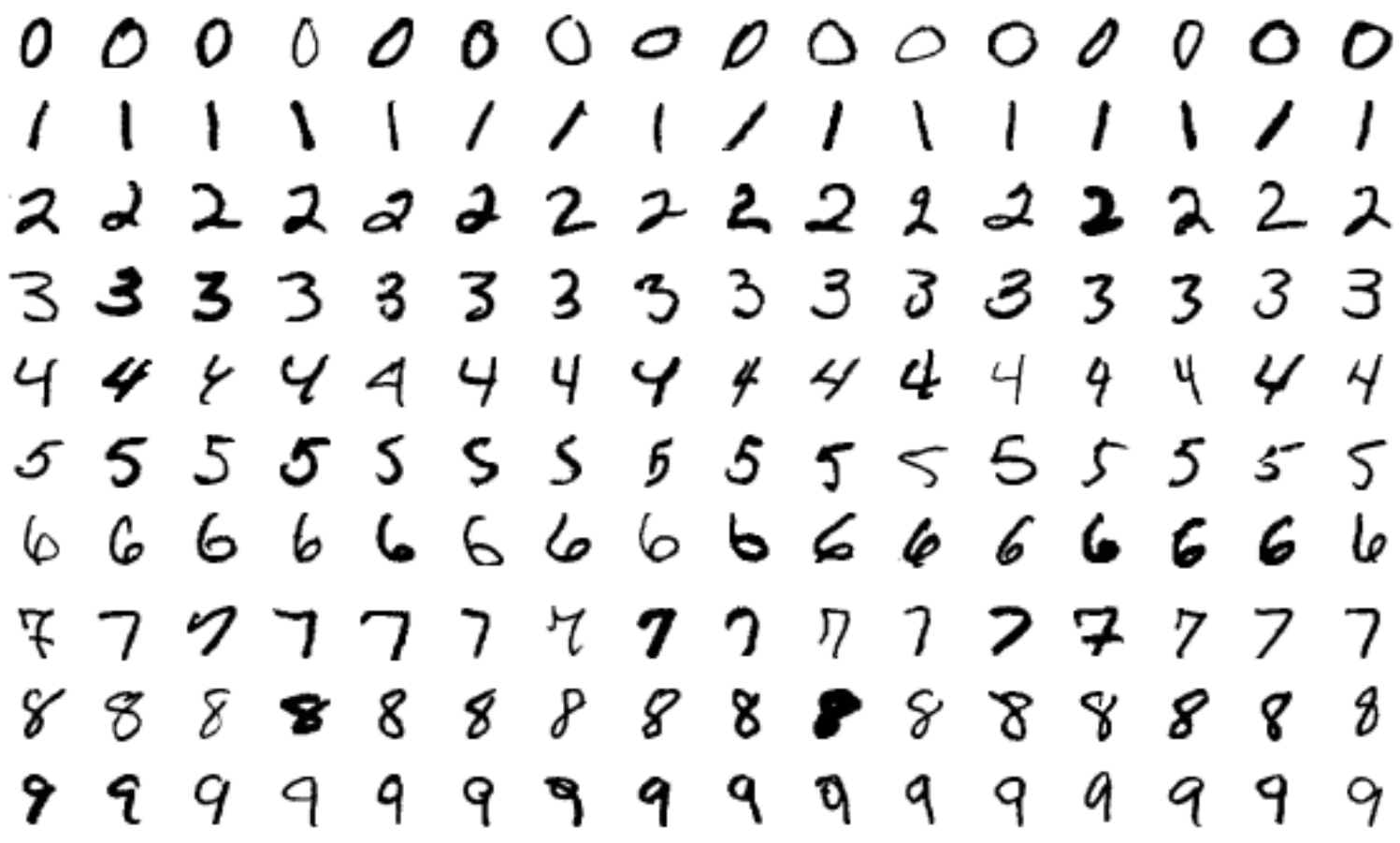

**Figure 5.1.** MNIST overview.

### 5.1.2 CIFAR-10

The CIFAR-10 [66] dataset contains 60,000 32x32 RGB images divided into ten classes, each with 6,000 images. There are 50,000 photos for training and 10,000 images for testing.

Each of the 10,000 photos in the dataset is separated into five training batches and one test batch. The test batch contains exactly 1,000 photos from each class, chosen at random. The remaining photographs are randomly distributed among the training batches, however certain training batches may contain more images from one class than the others. The training batches contain exactly 5,000 photos from each class between them. An overview can be found in Fig. 5.2.

### 5.1.3 CIFAR-100

This dataset [66] is similar to the CIFAR-10 dataset, but it comprises 100 classes, each with 600 photos. Each class has 500 training photos and 100 test images. The CIFAR-100's 100 classes are divided into 20 superclasses. Each image has a "fine" and a "coarse" designation, respectively the class to which it belongs and the superclass to which it belongs.

[0]Image taken from the site: https://www.cs.toronto.edu/ kriz/cifar.html

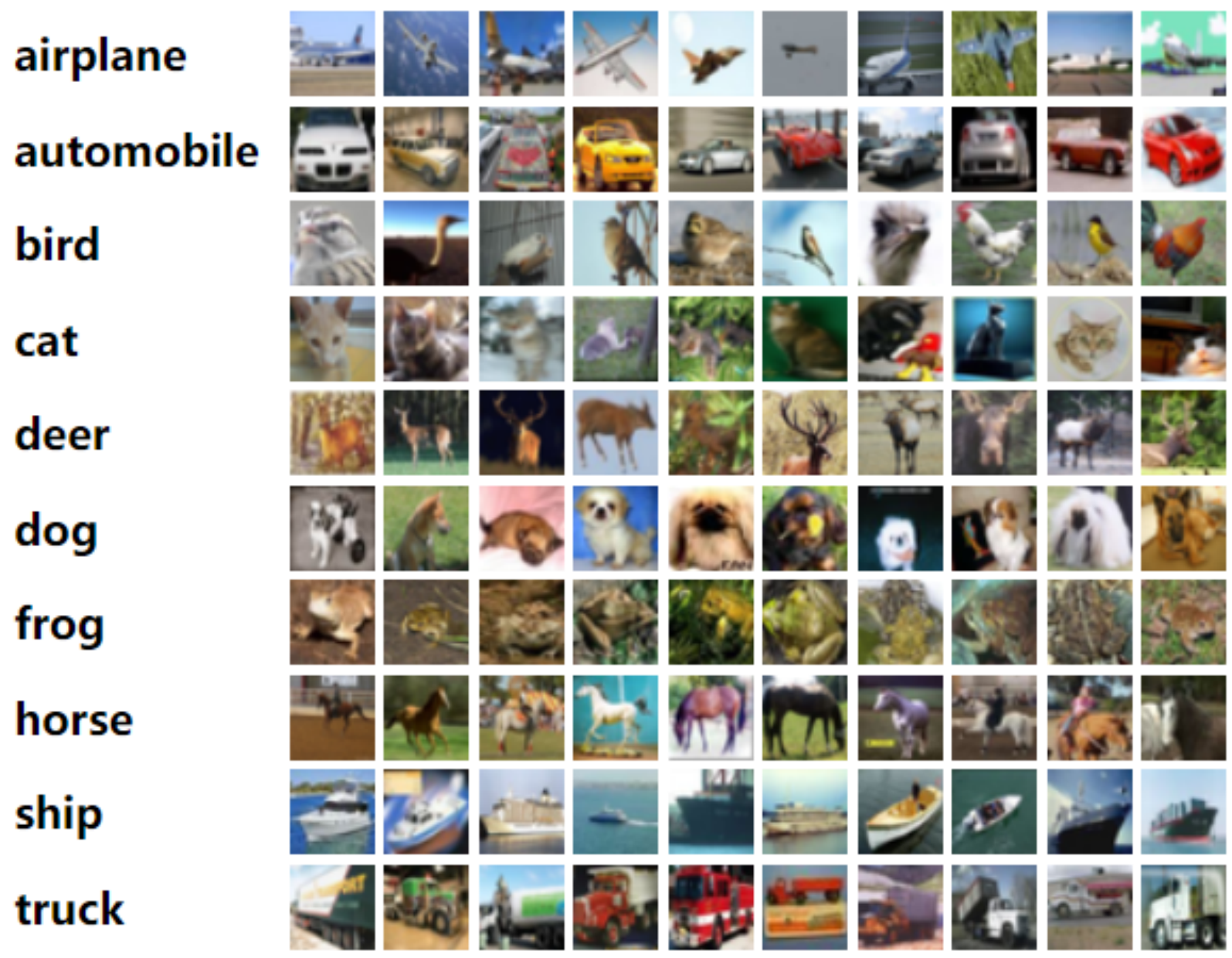

**Figure 5.2.** CIFAR-10 overview.

## 5.2 Implementation details

### 5.2.1 Deep learning framework: PyTorch

PyTorch is an open source deep learning framework based on the Torch library, largely created by Facebook's AI Research Division (FAIR) for applications such as computer vision and natural language processing. It's an open-source software distributed under the modified BSD license. PyTorch also provides a C++ interface, albeit the Python interface is more refined and the primary focus of development.

### 5.2.2 Cloud training: Google Colab

Google Colab is a cloud-based Jupyter notebook environment that is free to use. Most significantly, there is no setup required, and the notebooks you create can be modified simultaneously by your team members, much like documents in Google Docs. Many common machine and deep learning libraries are supported by Colab and can be quickly loaded into the notebook. Colab Pro features the following resources:

- CPU: Intel(R) Xeon(R) - 2 core 1.3 GHz;
- RAM: 13 GB;
- GPU: Nvidia K80 or T4, depending on the availability at the moment;
- GPU RAM: 12GB or 16GB, depending on the availability at the moment;

### 5.2.3 Meta datastore: Neptune.ai

Neptune is a meta datastore for research and production teams who do several experiments. It provides a central location for all metadata generated during the machine learning lifecycle to be logged, stored, displayed, organized, compared, and queried. Individuals and organizations utilize Neptune for experiment recording and model registry in order to maintain control over their research and development.

## 5.3 Results

A good evaluation phase also needs to compare the obtained results against other state of the art approaches; as a result, the proposed method was compared to the existing literature in efficient image classification task. For comparisons, the literature employs the metric top-1 accuracy on validation: top-1 accuracy is the probability of correctly classify a sample from the validation set.

Table 5.1 shows the results on CIFAR-10, the more used benchmark dataset in this field.

The following are the reasons for training speed improvements and energy savings in the Adam optimizer setting: $\times 4$ for 8bit training [67] + $\times 1.3$ for cyclic precision [63] and $\times 2.36$ for shorter training (254 epochs instead of 600).

Meanwhile, in the SGD configuration the speed up is composed of: $\times 4$ for 8bit training + $\times 1.3$ for cyclic precision [63].

The memory saving in both configuration is due to 8bit training.

In the final tests, the configurations are set as:

- Adam optimizer, no weight decay, initial learning rate of 0.0025 with a cosine annealing learning rate scheduler;

- SGD optimizer, initial learning rate of 0.1 with a cosine annealing learning rate scheduler;

| Full Precision Base CNN | | | | | |
|---|---|---|---|---|---|
| | | Inference | | Training | |
| Name | Acc(%) | Speed-up | Memory save | Speed-up | Memory save |
| ResNet (2020d) [68] | **93.0** | x1 | x1 | x1 | x1 |
| BNN Accuracy Performance | | | | | |
| | | Inference | | Training | |
| Name | Acc(%) | Speed-up | Memory save | Speed-up | Memory save |
| XOR-Net (2016) [3] | 90.21 | **x58** | **x32** | x1 | x1 |
| Bi-Real-Net (2018) [69] | 89.21 | **x58** | **x32** | x1 | x1 |
| Main/Sub (2019) [70] | 86.39 | **x58** | **x32** | x1 | x1 |
| IR-Net (2020c) [71] | 91.5 | **x58** | **x32** | x1 | x1 |
| RBNN (2020) [72] | 92.2 | **x58** | **x32** | x1 | x1 |
| ReCU (2021) [61] | 92.8 | **x58** | **x32** | x1 | x1 |
| Proposed (Adam) | 90.6 | **x58** | **x32** | **x12.3** | **x4** |
| Proposed (SGD) | 92.09 | **x58** | **x32** | x5.2 | **x4** |

**Table 5.1.** Experimental comparison on CIFAR-10 on validation top-1 accuracy. Bold values are column-wise better.

CIFAR-100 and MNIST have never been used in literature for benchmarking binary neural networks, but some tests made by the candidate using full-precision ResNet and ReCU can be found in Tables 5.2 and 5.3.

## 5.4 Discussion

This chapter details the extensive tests that were carried out. Following a brief overview of the training/benchmark dataset, the implementation details for the proposed architecture were presented. Information about the machine on which the algorithms were run were presented, as well as all of the hyper-parameters used to make the architecture handle binary neural network and low precision training without significant accuracy loss.

| Full Precision Base CNN | | | | | |
|---|---|---|---|---|---|
| | | Inference | | Training | |
| Name | Acc(%) | Speed-up | Memory save | Speed-up | Memory save |
| ResNet (2020d) [68] | **75.72** [73] | x1 | x1 | x1 | x1 |
| **BNN Accuracy Performance** | | | | | |
| | | Inference | | Training | |
| Name | Acc(%) | Speed-up | Memory save | Speed-up | Memory save |
| ReCU (2021) [61] | 72.25 | **x58** | **x32** | x1 | x1 |
| Proposed (Adam) | 68.63 | **x58** | **x32** | **x12.3** | **x4** |
| Proposed (SGD) | 70.46 | **x58** | **x32** | **x5.3** | **x4** |

**Table 5.2.** Experimental comparison on CIFAR-100 on validation top-1 accuracy. Bold values are column-wise better.

| Full Precision Base CNN | | | | | |
|---|---|---|---|---|---|
| | | Inference | | Training | |
| Name | Acc(%) | Speed-up | Memory save | Speed-up | Memory save |
| ResNet (2020d)* [68] | **97.9** [74] | x1 | x1 | x1 | x1 |
| BNN Accuracy Performance | | | | | |
| | | Inference | | Training | |
| Name | Acc(%) | Speed-up | Memory save | Speed-up | Memory save |
| Our (SGD)* | 97.6 | **x58** | **x32** | **x5.3** | **x4** |

**Table 5.3.** Experimental comparison on MNIST on validation top-1 accuracy. Bold values are column-wise better.
*: training without augmentation.

The enhancements introduced in this work can be divided into two categories: improvements to increase accuracy (4.2.2, 4.2.3, 4.2.4, 4.2.7), and improvements to increase efficiency (4.2.5, 4.2.6).

Finally, a state of the art comparison was shown, obtained on the available binary neural networks works. As proved, the obtained scores demonstrated how the devised method outperforms the other methods in terms of efficiency while losing a small amount in top-1 accuracy.

It is worth noting that, while changes in precision during training affect validation top-1 accuracy, but they appear to have smaller effects on training loss, as can be seen in Fig. 5.3.

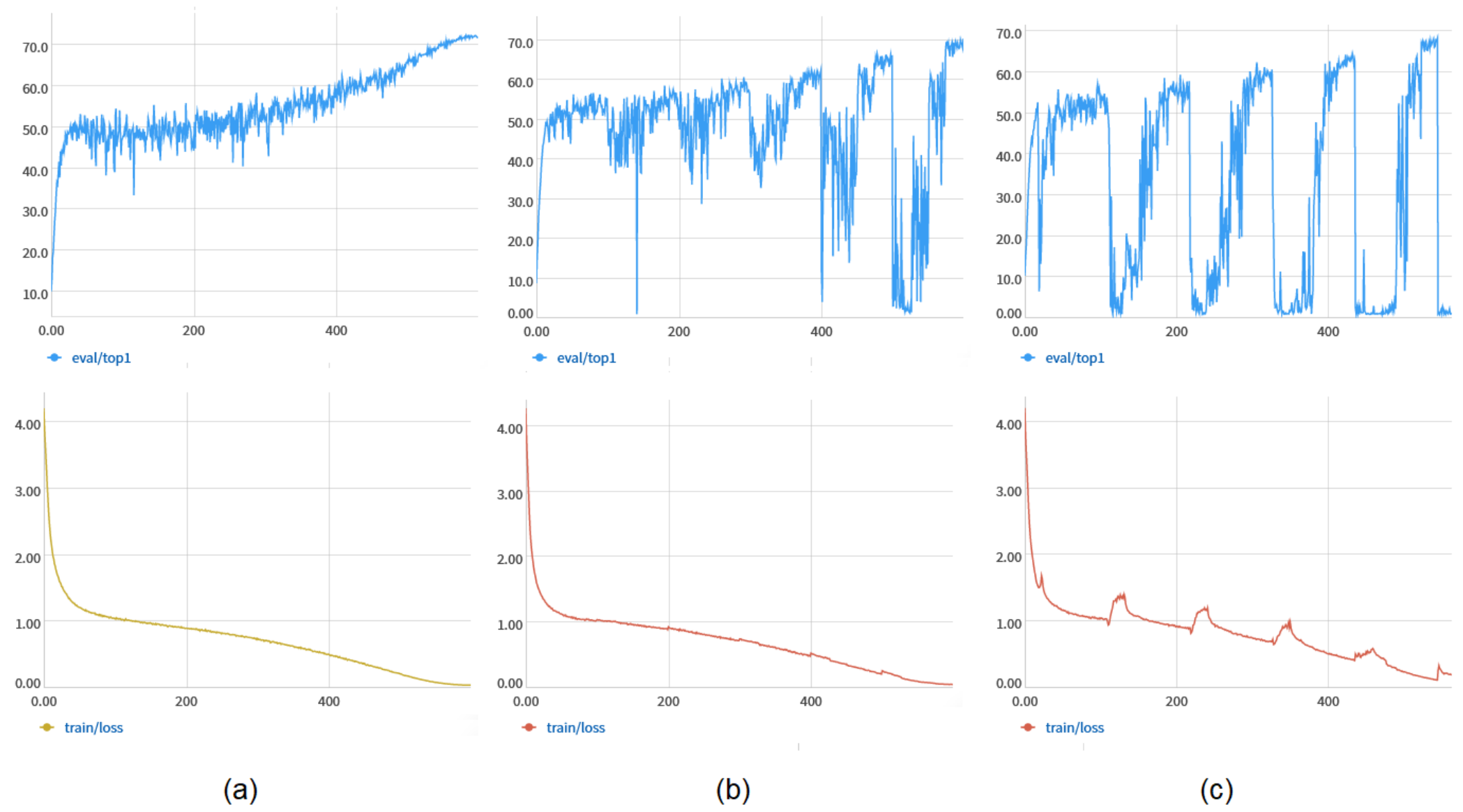

**Figure 5.3.** Benchmark on CIFAR-100: (a) ReCU's validation top-1 accuracy and train loss (b) Proposed with SGD's validation top-1 accuracy and train loss (c) Proposed with Adam's validation top-1 accuracy and train loss

# Chapter 6

# Conclusion

In this thesis are presented an original combination of tricks and improvements on a binary neural network in order to realize an original and much more efficient training.

An exhaustive experimental phase was performed on CIFAR-10 (the only benchmark common on all binary neural network solutions), CIFAR-100 and MNIST. The obtained results show how the proposed solution outperforms in terms of efficiency the works in this topic.

As a future development, interesting topics concerning the investigation of new feasible improvements to make the training more accurate and less expensive may be studied.

To reply to the question that's the thesis title: it is possible to train binary neural networks without floats, and it's can be very efficient.

# Appendices

# .1 Naive perturbation evolutionary's step

**procedure** NAIVE PERTURBATION EVOLUTIONARY'S STEP($N$ = starting network, $p$ = flipping probability )
    $NewNetwork = clone(N)$
    **for** each weight $w$ in $NewNetwork$ **do**
        flip $w$ with probability $p$
    **if** $evaluate(N) > evaluate(NewNetwork)$ **then**
        Return $N$
    **else**
        Return $NewNetwork$

# .2 Evolutionary algorithm's step

**procedure** GENETIC ALGORITHM'S STEP($elite$ = starting elite, $p$ = flipping probability, $children$ = number of children or each elite member, $elitesize$ = number of the networks in outputs )
    $population = Empty$
    **for** each Network $N$ in $elite$ **do**
        $newchildren = clone(N)$
        **for** each weight $w$ in $newchildren$ **do**
            flip $w$ with probability $p$
        append $newchildren$ to $population$
    $population = SortByScore(population)$# the score is a linear combination current accuracy and ancestors's accuracy
    Return $population[: population]$

# .3 Counting error evolutionary algorithm's step

**procedure** COUNTING ERROR EVOLUTIONARY ALGORITHM'S STEP($net$ = Starting network, $p$ = flipping probability, $children$ = number of children, $batchX$ = inputs, $batchY$=truth)

  $prediction = net(batchX)$

  set *wrong* the weights in the last layer that make $prediction \neq batchY$

  set *wrong* the input nodes in the last layer that have the majority of weights set to *wrong*

  **for** each layer $l$ (starting from the second-last going to the first) in $net$ **do**

    set *wrong* the weights in layer $l$ if connected to output node set *wrong*

    set *wrong* the input nodes in layer $l$ if the majority of the weights connected to them are set "wrong"

  $population = Empty$

  **for** each $n$ in range($children$) **do**

    $child = clone(net)$

    flip the weights of child set *wrong* with probability $p$

    append *child* to *population*

  flip the $w$ with probability $p$

  $population = SortByEvaluation(population)$

  Return $population[0]$